\documentclass[12pt,letterpaper]{article}
\usepackage{amssymb,amsmath}
\usepackage{mathtools}
\usepackage{amsthm}
\usepackage{bbm}
\usepackage{graphicx}
\usepackage{enumerate}
\usepackage{caption}
\usepackage{natbib}
\usepackage{url} 
\usepackage{bm}
\usepackage{color}
\usepackage{silence} 
\DeclareMathOperator*{\median}{\mbox{median}}

\DeclareMathOperator*{\argmax}{\mbox{argmax}}

\newcommand{\bv}{\boldsymbol v}
\newcommand{\bx}{\boldsymbol x}

\newcommand{\by}{\boldsymbol y}
\newcommand{\bz}{\boldsymbol z}

\newcommand{\balpha}{\boldsymbol \alpha}
\newcommand{\bbeta}{\boldsymbol \beta}

\newcommand{\bhSigma}{\boldsymbol{\widehat{\Sigma}}}

\newcommand{\hp}{\hat{p}}
\newcommand{\tp}{\tilde{p}}

\newcommand{\PAC}{\mbox{PAC}}
\newcommand{\farness}{\mbox{farness}}

\newcommand{\YJl}{h_{\lambda}}

\definecolor{blue}{RGB}{0,0,255}
\definecolor{red}{RGB}{255,0,0}

\begin{document}

\def\spacingset#1{\renewcommand{\baselinestretch}%
{#1}\small\normalsize} \spacingset{1}


\newcommand{\blind}{0}

\if0\blind
{
  \title{\bf Silhouettes and quasi residual
	           plots for neural nets and tree-based 
						 classifiers}
  \author{Jakob Raymaekers and Peter J. 
	  Rousseeuw \vspace{.3cm} \\
		Section of Statistics and Data Science,\\
		Department of Mathematics, KU Leuven, Belgium}
	\date{February 25, 2022}
  \maketitle
} \fi

\if1\blind
{
  \phantom{abc}
  \vskip2.0cm
  \begin{center}
	  {\LARGE\bf Silhouettes and quasi residual
	           plots for neural nets and tree-based 
						 classifiers}
	\vskip1.5cm
	{\large February 25, 2022}
	\vskip2cm
	\end{center}
} \fi

\begin{abstract}
Classification by neural nets and by tree-based 
methods are powerful tools of machine learning.
There exist interesting visualizations of the 
inner workings of these and other classifiers.
Here we pursue a different goal, which is to
visualize the {\it cases} being classified, 
either in training data or in test data.
An important aspect is whether a case has been
classified to its given class (label) or whether
the classifier wants to assign it to different 
class. 
This is reflected in the (conditional and
posterior) {\it probability of the alternative 
class} (PAC). A high PAC indicates label bias, 
i.e. the possibility that the case was mislabeled.
The PAC is used to construct a {\it silhouette plot}
which is similar in spirit to the silhouette plot
for cluster analysis 
\citep{Rousseeuw:Silhouettes}.
The average silhouette width can be used to 
compare different classifications of the same 
dataset.
We will also draw {\it quasi residual plots}
of the PAC versus a data feature, which may lead 
to more insight in the data.
One of these data features is how far each case
lies from its given class.
The graphical displays are illustrated and
interpreted on data sets containing 
images, mixed features, and tweets.
\end{abstract}


\spacingset{1.45} 

\section{Introduction} \label{sec:intro}
  
Classification by neural nets and by tree-based 
methods are powerful tools of machine learning.
\cite{Hastie:EOSL} give a broad overview
of classification methods.
There exist interesting visualizations of the 
inner workings of classification by neural nets;
see, e.g., \cite{Shahroudnejad} and the 
references cited therein.
Classification trees such as those produced
by CART \citep{CART} and the corresponding R 
package \texttt{rpart} \citep{rpart} can plot 
the tree and list its decision rules, which makes 
the classification explainable.
The random forest classifier \citep{Breiman:RF}
can be understood as an ensemble of such trees.

In this paper we propose visualizations of 
the {\it cases} being classified, rather than 
the mechanism of the classifier.
We are convinced that visualizing the objects 
being classified is equally useful, and can 
reveal different and very relevant aspects of 
the classification task.
The visualization of cases is well-established
for regression tasks, with e.g. residual plots,
whereas it is lagging behind for classification.
The purpose of the new visualizations is to draw 
our attention to interesting aspects of the data 
that we might not have known otherwise or did 
not expect. Examples are the strength of the
classification per label, noticing patterns in
subsets of the data, detecting mislabeled 
instances, and discovering overlap between classes. 
Interpreting such clues may provide insight into 
the structure and quality of the data.
The graphical displays can also reveal
underlying causes of misclassifications,
telling us something about the appropriateness 
of the classifier. 
This will be illustrated in various examples 
throughout the paper.

In classification, a key concept is 
the conviction with which an observation is 
assigned to its own class or a different class. 
This information is 
captured by the probability of the alternative 
class (PAC) which is fundamental to our 
visualizations.
We use the PAC to construct a 
{\it silhouette plot} which is similar in spirit 
to the silhouette plot for unsupervised 
classification \citep{Rousseeuw:Silhouettes}.
The average silhouette width (on test data or
cross-validated) can be used to compare different 
classifiers applied to the same dataset.
We will also draw {\it quasi residual plots}
of the PAC versus a data feature. Patterns in
such plots may reveal interesting trends, and
may help unearth factors explaining why some 
cases are easier to classify correctly than 
others, potentially informing model choice.  

In subsection \ref{sec:NN} we will focus on neural
nets, and analyze the results of a classification
of images from 10 categories.
Subsection \ref{sec:rpart} applies the general
principles to classification by CART, and
subsection \ref{sec:RF} does the same for random
forests, each illustrated on a well-known dataset.
Section \ref{sec:conclusions} concludes and
describes the available software.

\section{Silhouette plots for classification} 
\label{sec:silh}

The silhouette plot of \cite{Rousseeuw:Silhouettes}
is a graphical display of a clustering (unsupervised
classification) in $k$ clusters.
The silhouette width 
\begin{equation} \label{eq:si}
  s(i) := \frac{b(i) - a(i)}
	       {\max \big(a(i),b(i)\big)} 
\end{equation}
describes how well case $i$ is clustered.
Here $a(i)$ is the average interpoint dissimilarity 
of case $i$ to the members of its own cluster.
In contrast, $b(i)$ is the smallest average 
dissimilarity of case $i$ to a non-self cluster.
That cluster can be considered the `best alternative'
cluster for case $i$.
From \eqref{eq:si} we see that $s(i)$ is between
-1 and 1.
When $s(i)$ is high (close to 1) it means that case 
$i$ has much more in common with its own cluster 
than with any other cluster, so it was clustered well.
On the other hand, an $s(i)$ close to -1 means that
case $i$ would much prefer to be assigned to its
best alternative cluster.

\begin{figure}[!ht]
\centering
\vspace{0.2cm}
\includegraphics[width = \textwidth]
  {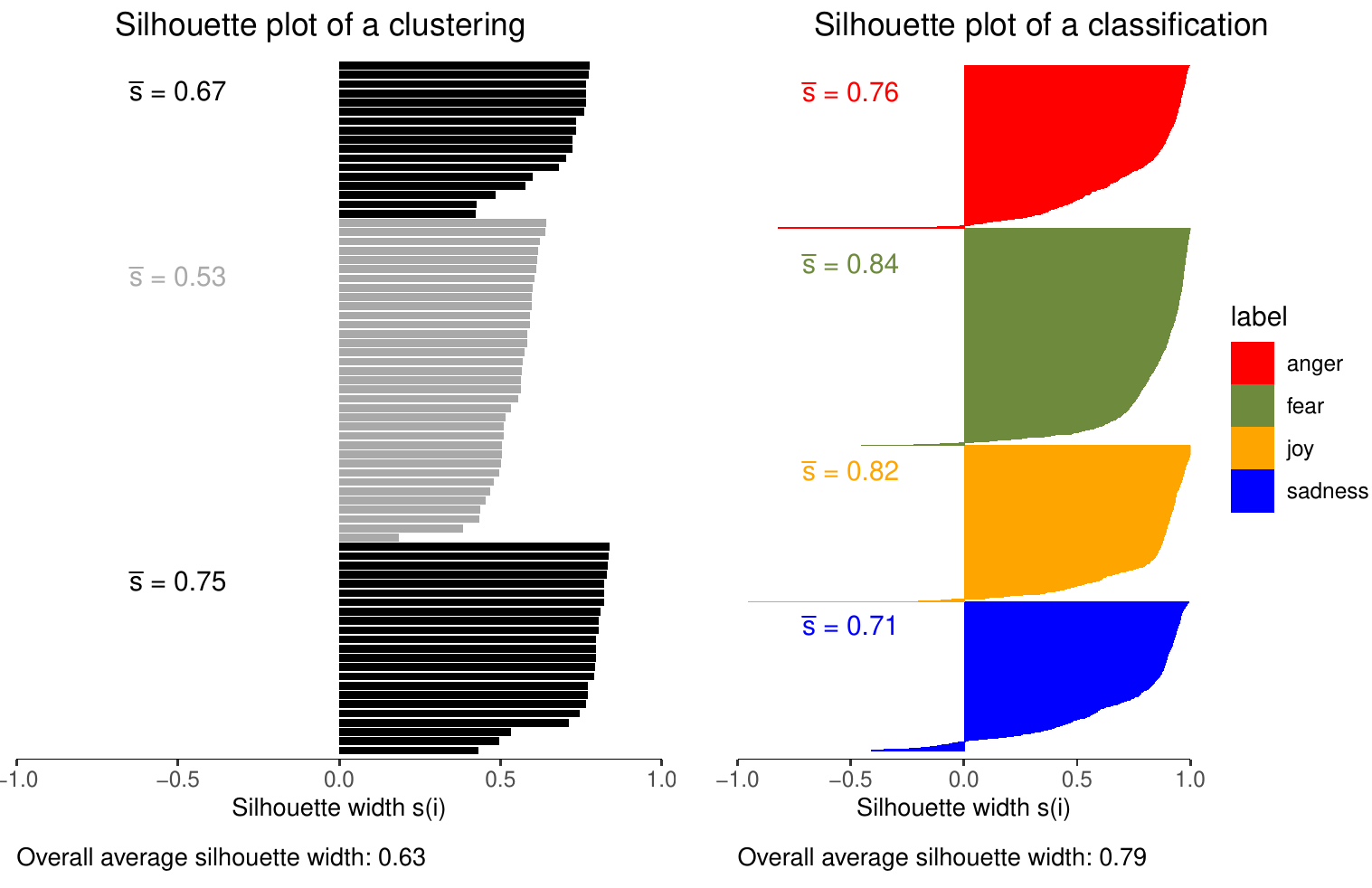}	
\caption{Silhouette plots of (left) a 
  partitioning of a toy dataset in three 
	clusters, and (right) a classification 
	with four classes.}
\label{fig:silh1}
\end{figure}

The silhouette plot displays the s(i) values, ranked
in decreasing order in each cluster.
The silhouette of a cluster reflects how well its 
members are clustered.
The left panel of Figure \ref{fig:silh1} shows
the silhouettes of a partition with $k=3$ of a toy
dataset.
The $s(i)$ are on the horizontal axis.
For instance, in the second cluster from the top
the cases range from well-clustered (high $s(i)$)
to poorly clustered (low $s(i)$).
The silhouette of the third cluster is the widest,
and indeed the average of its $s(i)$, shown on the
left as $\bar{s} = 0.75$, is the highest.
The {\it overall average silhouette width} of
$0.63$ listed at the bottom is the average $s(i)$ 
over all cases $i$ in the data set.
Usually the number of clusters $k$ is not given in
advance, and then one often selects the value of 
$k$ that makes the overall average silhouette 
width the highest. 

In our setting of supervised classification, the
situation is somewhat different. 
Not all classifiers are based on interpoint
dissimilarities; in particular, neural nets
combine the outputs of cells in neurons, and 
tree-based classifiers use variable splits.
We denote a class (label, group) by the
letter $g$, with $g = 1,\ldots,G$.
Consider a case $i=1,\ldots,n$ in the training 
set or a test set. 
Typically, a classifier will provide posterior 
probabilities $\hp(i,g)$ of object $i$ belonging 
to each of the classes $g$, with 
$\sum_{g=1}^G{\hp(i,g)} = 1$ for each $i$.
The object $i$ is then classified according to the
{\it maximum a posteriori} rule
\begin{equation} \label{eq:MAP}
  \mbox{assign object }\; i \; 
	\mbox { to class }\;\;
	\argmax_{g=1,\ldots,G}\; \hp(i,g)\;.
\end{equation}
Now consider the object $i$ with its known given 
label $g_i$\,.
In analogy with the $s(i)$ above, we want to 
measure to what extent the given label $g_i$ 
agrees with the classification of $i$.
For this purpose we define the highest $\hp(i,g)$ 
attained by a class {\it different from} $g_i$ as
\begin{equation}\label{eq:altclass}
   \tp(i) := \max\{\hp(i,g)\,;\,g \neq g_i\}\;.
\end{equation}
The class attaining this maximum can be seen as 
the best alternative class, so it plays the
same role as the class yielding $b(i)$ in
clustering.
If $\hp(i,g_i) > \tp(i)$ it follows
that $g_i$ attains the overall highest value of 
$\hp(i,g)$, so the classifier agrees with the given 
class $g_i$\,. 
On the other hand, if $\hp(i,g_i) < \tp(i)$
the classifier will not assign object $i$ to 
class $g_i$\,. 

We now compute the conditional posterior 
probability of the best alternative class when 
comparing it with the given class $g_i$ as 
\begin{equation}\label{eq:PAC}
	\PAC(i) = \frac{\tp(i)}{\hp(i,g_i) + \tp(i)}\;\;.
\end{equation}
The abbreviation PAC stands for Probability of
the Alternative Class.
It always lies between 0 and 1, and smaller 
values are better than larger values.
When $\PAC(i) < 0.5$ the classifier does predict 
the given class $g_i$\,, whereas $\PAC(i) > 0.5$ 
indicates that the best alternative class 
outperforms $g_i$ in the eyes of the classifier.
$\PAC(i) \approx 0$ indicates that the given class
fits very well, and $\PAC(i) \approx 1$ means the
given class fits very badly.
The PAC can be seen as a continuous alternative
to the more crude distinction between "correctly
classified" and "misclassified" that is 
used in the misclassification rate.

In order to draw the silhouette plot of a 
classification, we put
\begin{equation} \label{eq:silclassif}
  s(i)\;:=\; 1-2\,\PAC(i)\;.
\end{equation}
Like~\eqref{eq:si} this $s(i)$ again ranges from
-1 to 1, with high values reflecting that the 
given class of case $i$ fits very well, and
negative values indicating that the given class
fits less well than the best alternative class.
The actual silhouette plot is then drawn as
before, for example in the right panel of 
Figure~\ref{fig:silh1} with $G=4$ classes
shown in different colors. 
The fact that the $s(i)$ have a continuous 
range allows us to see finer detail than if 
we would only display whether a case is 
classified correctly or not.

The data and the classifier leading to this
plot will be explained in 
subsection~\ref{sec:RF}, but the display
alone already tells us a lot. 
The silhouettes have unequal heights, which
are proportional to the number of cases
in each given class.
Each class has several cases with high $s(i)$, 
that are predicted strongly in it,
but also some cases with negative $s(i)$, 
which the classifier predicts in a different 
class.
The overall silhouette width is 0.79\,.
Class `fear' has the highest average
silhouette width ($\bar{s} = 0.84$), 
similar to that of class `joy' (0.82), and 
followed by classes `anger' (0.76) and 
`sadness' (0.71). 
This indicates that classes `fear' and `joy' 
are fit best by this particular classifier.

In supervised classification the number of 
classes $G$ is known in advance, so the overall 
average silhouette width cannot play the same 
role as in cluster analysis, where it is used 
to select the number of clusters. 
But when different classifiers are applied to 
the same dataset, it measures 
the quality of each classification, so one 
could select the classification with the 
highest overall average silhouette width.
	
\section{Quasi residual plots} 
\label{sec:quasi}
	
Another graphical display is obtained by plotting
the $\PAC$ versus a relevant data variable.
This is not unlike plotting the absolute residuals
in regression, since small values of $\PAC(i)$
indicate that the model fits the data point nearly
perfectly, whereas a high $\PAC(i)$ alerts us to
a poorly fitted data point.
We call it a {\it quasi residual plot} because
of this analogy.
The data feature on the horizontal axis does not
have to be part of the classification model,
and it could also be a quantity derived from the 
data features such as a principal component or
a prediction, or just the index $i$ of the data 
point if the data were recorded sequentially.

\begin{figure}[!ht]
\centering
\includegraphics[width = 0.85\textwidth]
  {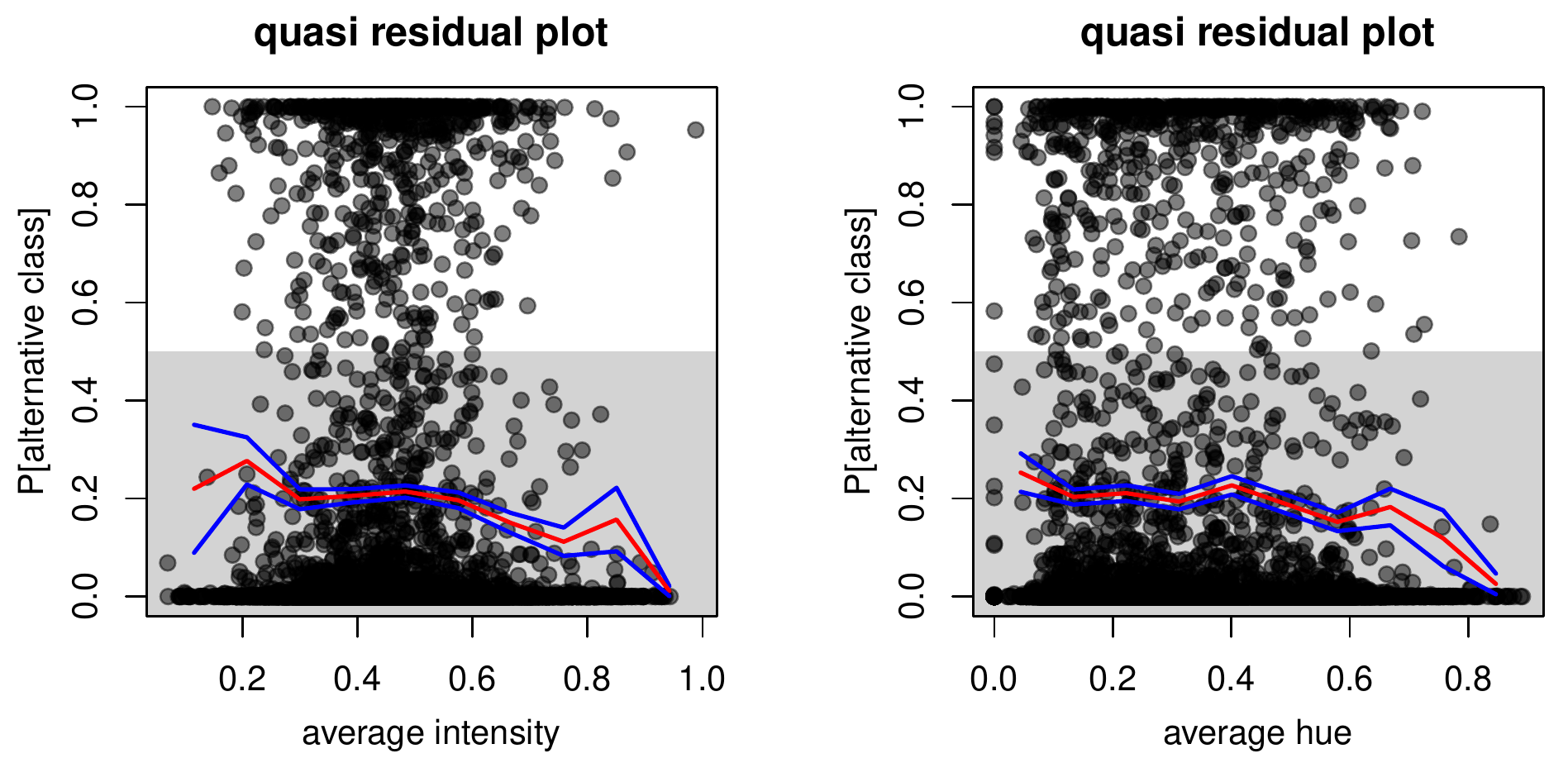}
\caption{Quasi residual plots of images versus
         their intensity and hue.}
\label{fig:qrp1}
\end{figure}	

Figure~\ref{fig:qrp1} shows two quasi residual 
plots. The data will be described in 
subsection~\ref{sec:NN}, and consists of 10,000
color images with $G=10$ classes.
The classifier has operated on the pixels of each
image, yielding the PAC on the vertical axis.
The variables on the horizontal axis were not
in the classification model. The left panel plots
the PAC versus the intensity of each image, which
was averaged over all pixels and the three colors
(red, green, and blue). 
Note that the bottom half of the plot has a 
light gray background. Points in this region have
$\PAC(i) < 0.5$, meaning that they are predicted
in their given class, whereas the classifier 
would put the points in the top half in a
different class.

Since the density on the
horizontal axis is far from uniform, three curves
were superimposed. The red curve is the average
PAC on 10 equispaced intervals, positioned in
the midpoint of each interval, after which the 
averages were connected by line segments. The blue
curves correspond to the average plus or minus
one standard error.
We see that higher intensities tend to yield
lower PAC, meaning that the classifier did a
better job on them. In the right hand panel we
see a similar effect in function of the hue of 
each image. 

A possible strategy is to record, 
collect, or construct a 
number of features and to run a regression 
method with the $\PAC$ as response variable. 
This may tell the user which factors affect 
the precision of the classifier. With that
information one could finetune the classifier, 
or select additional features for the 
classification.

\subsection{Class maps}

Class maps \citep{VCR} are quasi residual 
plots versus a feature reflecting how far each 
case is from its class.
This is based on some distance measure $D(i,g)$ 
of a case $i$ relative to a class $g$.
For each classifier in section 
\ref{sec:classifiers} we will specify an
appropriate measure $D(i,g)$.

Next we estimate the cumulative distribution 
function of $D(\bx,g)$ where $\bx$ is a random 
object generated from class $g$\,. 
The {\it farness} of the object $i$ to the class 
$g$ is then defined as
\begin{equation}\label{eq:farness}
 \mbox{\farness}(i,g) :=
   P[D(\bx,g) \leqslant D(i,g)]\;.
\end{equation}
Therefore $\farness(i,g)$ lies in the $[0,1]$ 
range, just like $\PAC(i)$.
The computation of \eqref{eq:farness} is
described in section A.1 
of the Supplementary Material.

The class map plots $\PAC(i)$ versus $\farness(i,g)$ 
for all cases $i$ with given label $g$.
The colors of the points are those of the predicted 
class. Points with high farness to {\it all}
classes are marked by a black border, as in
Figure 4.


\section{\mbox{Silhouette and quasi residual plots 
applied to classifiers}} \label{sec:classifiers}

The silhouette plot, quasi residual 
plot and class map can be drawn for training data
as well as labeled test data, with the same 
interpretation. But the motives for looking at
them are slightly different. 
Plots of the training data can help to assess 
whether the classifier is appropriate, discover
overlap between classes, and find mislabeled 
points so their labels can be corrected to improve 
the trained model.
Making plots of the test data can yield the 
same type of conclusions, but can also highlight 
aspects specific to the test data, such as 
differences between training and test data.
They can also help identify gaps in the training 
data. If a test image of a cat in the snow is 
classified as a dog because only dogs appeared 
in the snow in the training data, it might be good
to add images of cats in the snow to the training
data.

\subsection{Neural nets} 
\label{sec:NN}

In this section we illustrate the proposed
graphical displays in the setting of 
classification by a neural network.
Neural networks encompass a broad class of 
classifiers which are based on a structure with
an input layer, hidden layers, and an output layer, 
each consisting of a number of nodes. 
For a classification into $G$ groups based on 
$p$-variate data, the input layer has $p$ nodes,
each corresponding to one input variable, and the 
output layer has $G$ nodes, one for each class. 
The number and sizes of hidden layers 
and their connections determine the structure of 
the network and have to be fixed beforehand.

For an introduction to neural networks we refer to 
\cite{Hastie:EOSL}.
In its simplest form, a neural network classifier
has one hidden layer with $M$ nodes. 
Case $i$ is described by a $p$-dimensional vector 
$\bx_i$ of input variables.
The response is its given class $g_i$\,. 
This is encoded as a $G$-variate vector $\by_i$
which has 1 in the position $g_i$ and 0 in all
other positions.
(This is called `one-hot encoding'.)
We aim to approximate 
the response by a function $f$, that is,
$\by_i \approx f(\bx_i)$. 
The neural network will then create $M$ new 
features in the intermediate layer, given by
\begin{equation*}
  (\bz_i)_m = \sigma(\alpha_{0m}+\balpha_m'\bx_i) 
\;\;\; \mbox{ for } \;\; m = 1,\ldots,M.
\end{equation*}
For the activation function $\sigma$ one often
takes the rectified linear unit 
$\sigma(t) := \max(0,t)$. 
Next, $G$-variate vectors $\bv_i$ are obtained 
as linear combinations of the vectors $\bz_i$ by
\begin{equation} \label{eq:vi}
  (\bv_i)_g = \beta_{0g} + \bbeta_g'\bz_i  
  \;\;\; \mbox{ for } \;\; g = 1,\ldots,G.
\end{equation}
These vectors $\bv_i$ do not yet contain
probabilities. 
To that end one applies the $G$-variate 
{\it softmax} function $h$ given by
\begin{equation} \label{eq:softmax}
  (h(\bv_i))_g = \frac{e^{(\bv_i)_g}}
     {\sum_{j=1}^{G}{e^{(\bv_i)_j}}}\;\;.
\end{equation}		
The end result is the vector 
 $f(\bx_i) := h(\bv_i)$	
with positive entries. These can be seen as
posterior probabilities $\hp(i,g) := f(\bx_i)_g$\,. 
They indeed satisfy $\sum_{g=1}^G{\hp(i,g)} = 1$ 
by virtue of~\eqref{eq:softmax} in the final layer 
of the network.
The PAC can then be calculated 
from~\eqref{eq:altclass} and \eqref{eq:PAC}.

In practice most neural networks have multiple hidden
layers, that are chained to each other to allow more 
complex structures to be learned from the data. 
These hidden layers are connected in the same way as 
the layers above, i.e. by applying a nonlinear 
activation function on a linear combination of the 
outputs of the previous layer. Some of these layers
can have specialized connections depending on the 
classification task at hand.
For instance, for classifying images, convolutional 
neural networks (CNNs) are very popular. 
They incorporate `convolutional layers' and 
`pooling layers' that combine the information in 
nearby pixels (`local connectivity') to capture 
spatial information and reduce the dimension.
Neural networks are most commonly trained by
backpropagation, which allows gradient-based 
optimization of a loss function.
The model is good when the fitted vectors $f(\bx_i)$
are close to the response vectors $\by_i$\,.
Training can take long, but for a new case $\bx$
the prediction $f(\bx)$ is fast.

As an illustration we use the well-known
CIFAR-10 benchmark dataset.  
It consists of 60,000 color images with 
$32 \times 32$ pixels. They depict objects from 10 
classes, with 6000 images per class.
There are 50,000 training images and 10,000 test 
images. 

\begin{figure}[!ht]
\centering
\includegraphics[width=0.75\columnwidth]
  {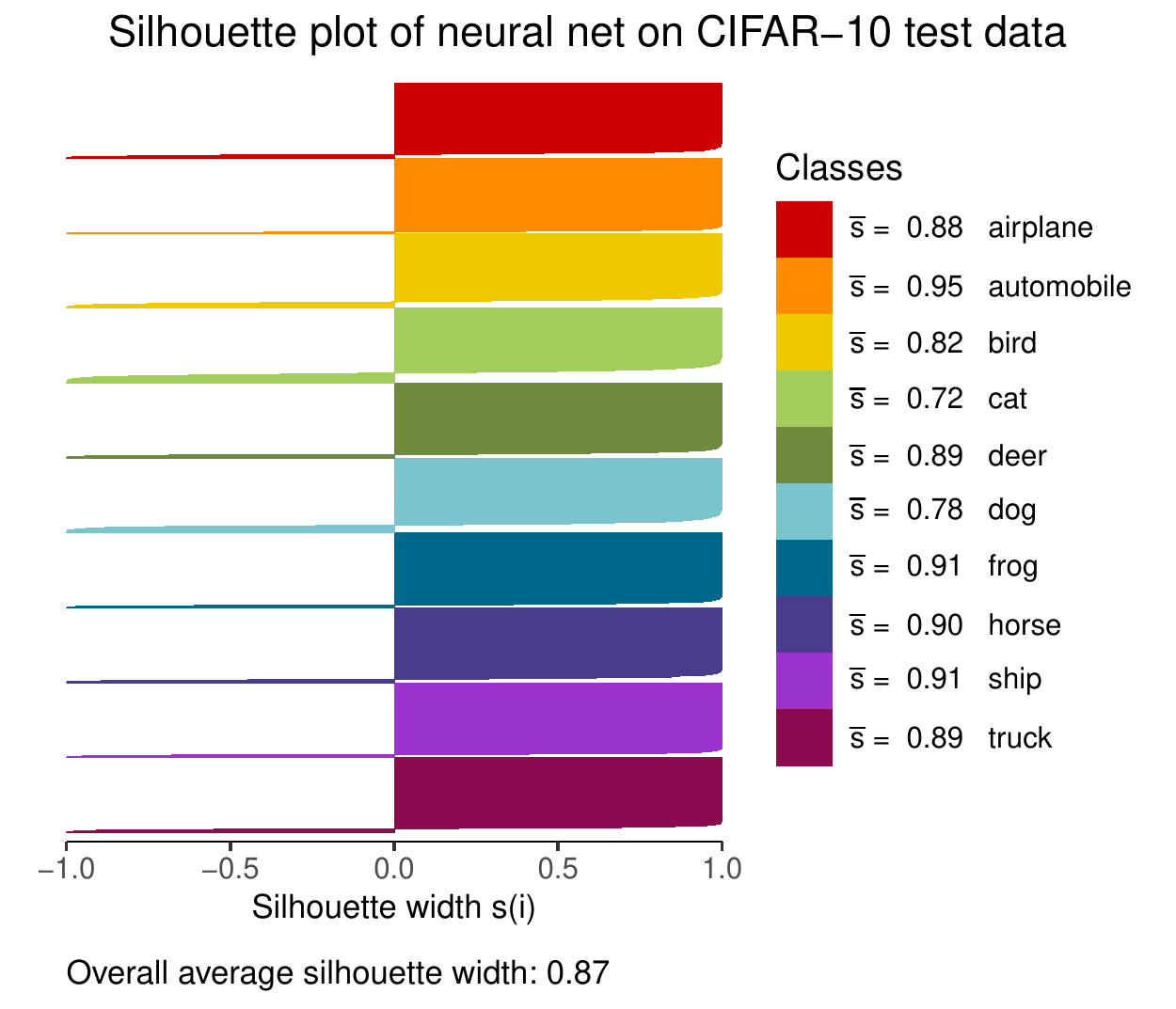} 
\caption{Silhouette plot on the CIFAR-10 test data.}
\label{fig:cif10_sil}
\end{figure}

The CIFAR-10 data have been classified by the 
residual neural network with 56 layers (ResNet-56) 
of \cite{he2016deep}. 
We have looked at the proposed graphical 
displays on the training data (not shown), but 
they are not very eventful because the model 
obtains a perfect classification on the training
data.
On the test data the accuracy is a realistic 94\%, 
allowing for more interesting visualization. 
Figure \ref{fig:cif10_sil} shows the
silhouette plot of the test data. 
With an overall average silhouette width of 0.87, 
we can conclude that the test data is classified 
quite well.
We also see clear differences between the classes. 
The class of automobiles has the highest average
silhouette width $\bar{s} = 0.95$, so the classifier
did best on this class. 
Animals seem to be harder to classify, with cats 
and dogs obtaining an average silhouette width 
$\bar{s}$ below 0.8. 
Also note that this classifier often had a
high conviction, with many $s(i)\approx 1$ (when 
classified correctly) or $s(i)\approx -1$ (when 
misclassified).

Now we turn to quasi residual plots. For image data,
it would not be natural to plot the PAC versus a 
single input feature, since this would 
correspond to the red, green or blue value in  
one pixel. Instead, we used the summary 
properties of intensity and hue of an image in 
Figure~\ref{fig:qrp1}, already shown in 
subsection~\ref{sec:quasi}.

For the class maps described in section
\ref{sec:quasi} we start by computing
the Mahalanobis distance $D(i,g)$ of each case
$i$ relative to each class $g$, given by
\begin{equation} \label{eq:MD}
 D(i,g) := \sqrt{(\bv_i - \bar{\bv}_g)'
   \bhSigma_g^{-1}(\bv_i - \bar{\bv}_g)}
\end{equation}
where $\bv_i$ is given by \eqref{eq:vi},
$\bar{\bv}_g$ is the average of all $\bv_j$ in
class $g$ in the training data, and 
$\bhSigma_g$ is their covariance matrix.
This requires that all the $G \times G$ matrices
$\bhSigma_g$ are nonsingular, which is typically 
the case when each class has many members 
compared to $G$. 
The resulting farness is then given by
\eqref{eq:farness}.

\begin{figure}[!ht]
\vspace{0.3cm}
\centering
\includegraphics[width = 0.6\textwidth]
  {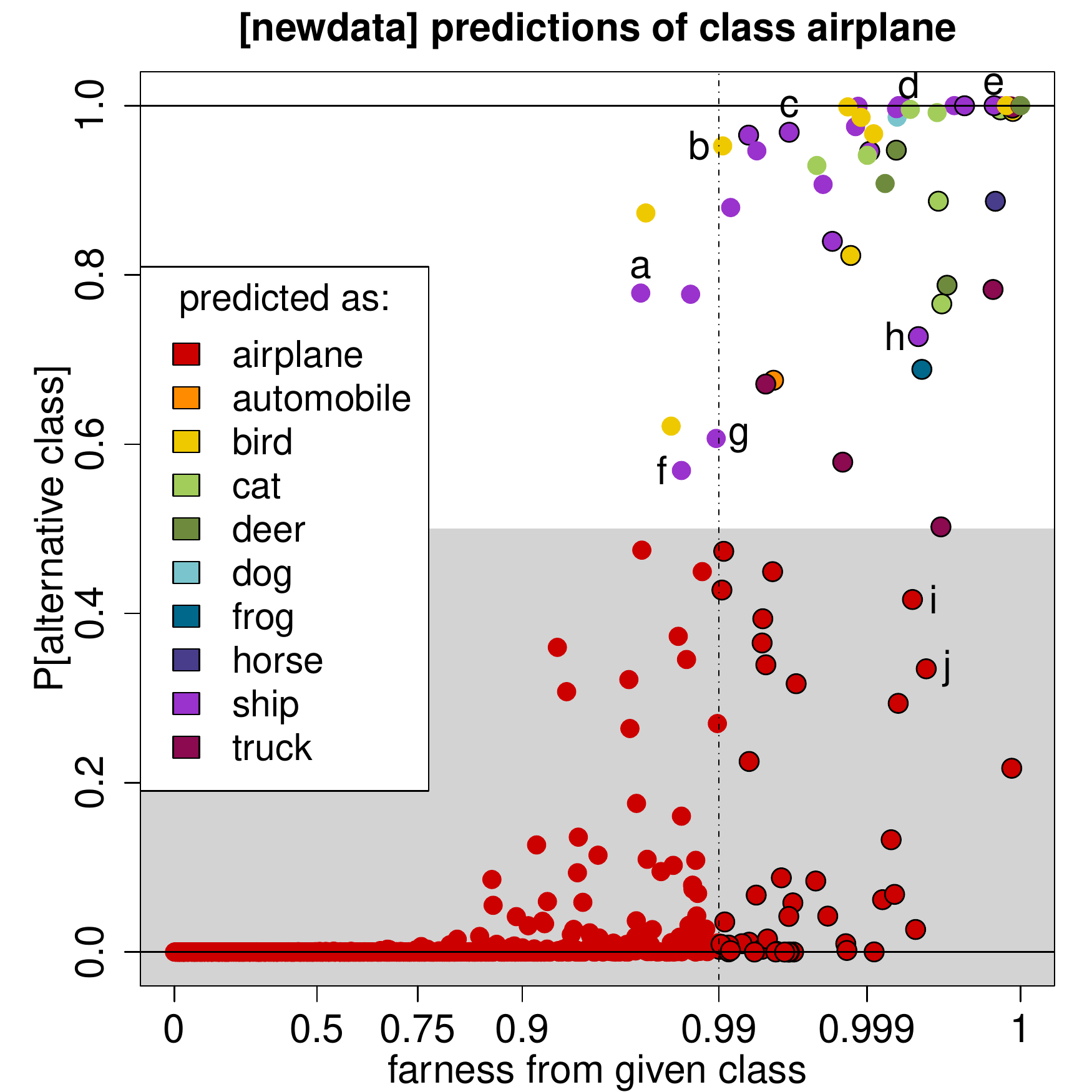}\\
\vspace{0.3cm}
\includegraphics[width = 0.8\textwidth]
  {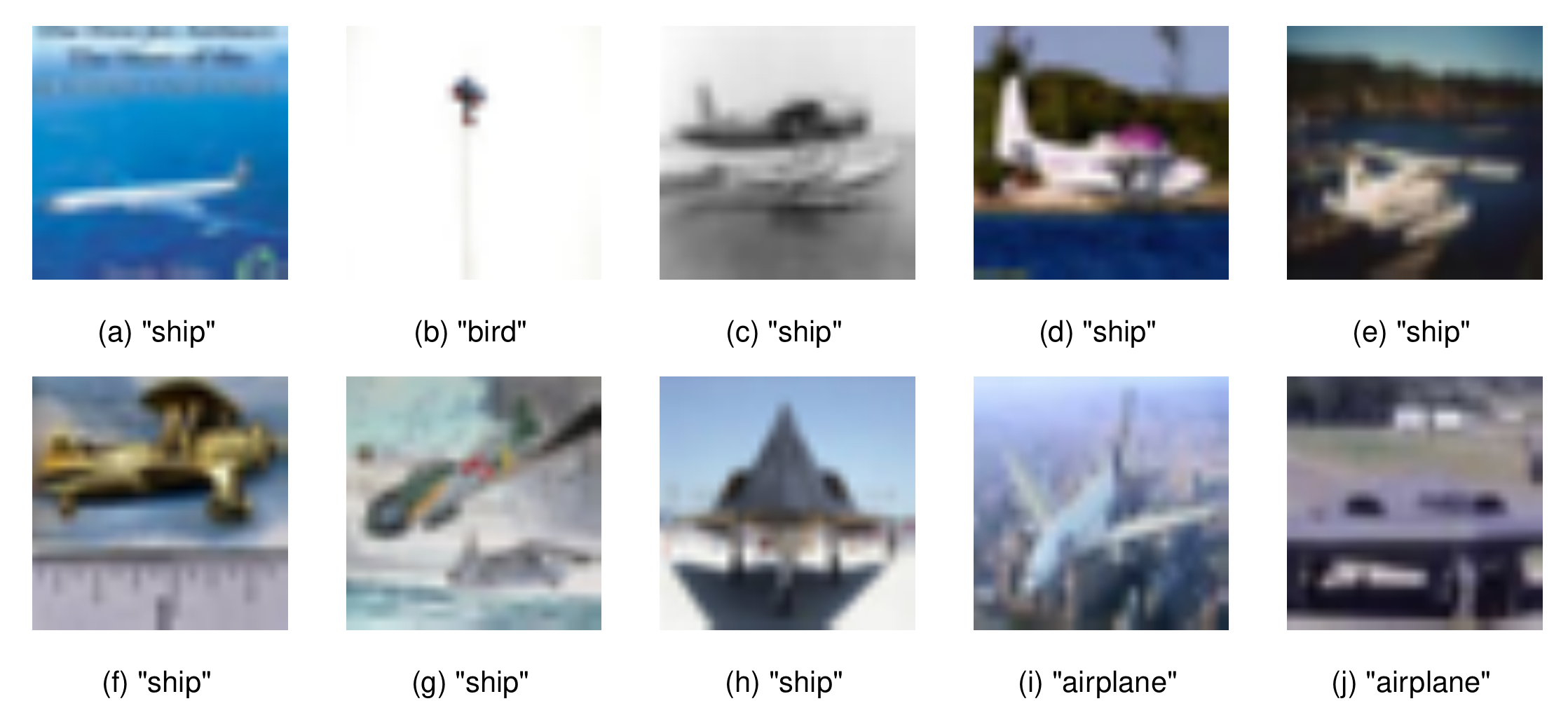}
\caption{Class map of the airplane class in the test 
 data, with the images corresponding to the marked 
 points.}
\label{fig:cifar_classmap_airplane}
\end{figure}

Figure~\ref{fig:cifar_classmap_airplane} is the 
class map of airplane images in the test data.
We see that most of the images get classified
correctly, as the majority of points have 
a PAC below 0.5 and are shown in red, the color
of this class. The misclassified cases are the
points with $\PAC(i)$ above 0.5, with many purple 
points being classified as ships and some yellow 
points as birds. These unusual cases stood 
out in the class map, and deserve to be looked at.
The images corresponding to the marked points 
are shown below the class map. Images 
\texttt{a}, \texttt{c}--\texttt{e} and \texttt{g}
assigned to class `ship' have water in them.
Three of these are seaplanes (\texttt{c}--\texttt{e}),
and \texttt{g} contains both an airplane and a ship.
Image \texttt{b} is classified as a bird, which
is not too surprising as it does look like one.
The object covers so few pixels that 
it is hard to classify, even for a human.
Finally, points \texttt{i} and \texttt{j} are
correctly classified as airplanes, but with high
farness. The first is an airplane photographed
from an unusual angle, with a city as background. 
The second has a strange shape, and could be a 
stealth plane.

The farness probabilities on the horizontal axis
are not equispaced: they are shown on the 
scale of quantiles of the standard gaussian 
distribution restricted to the interval [0,4].
This makes high farness values stand out more.
The vertical dashed line is at a cutoff value,
which can be chosen by the user and is 0.99 by
default. Cases which are far from every class in 
the data, that is with $\farness(i,g)$ above the
cutoff for all $g$, are called 
\textit{farness outliers} and plotted with a 
black border in the class map. Such cases do not 
lie well within any class, for example the images
\texttt{c}, \texttt{e}, and 
\texttt{h}--\texttt{j}.
Note that class maps are similar in spirit to 
the regression outlier maps of 
\cite{Rousseeuw:Diagnostic} and
\cite{Rousseeuw:MCDreg}, which plot 
residuals versus farness to the entire dataset.

\begin{figure}[!ht]
\vspace{0.3cm}
\centering
\includegraphics[width = 0.6\textwidth]
  {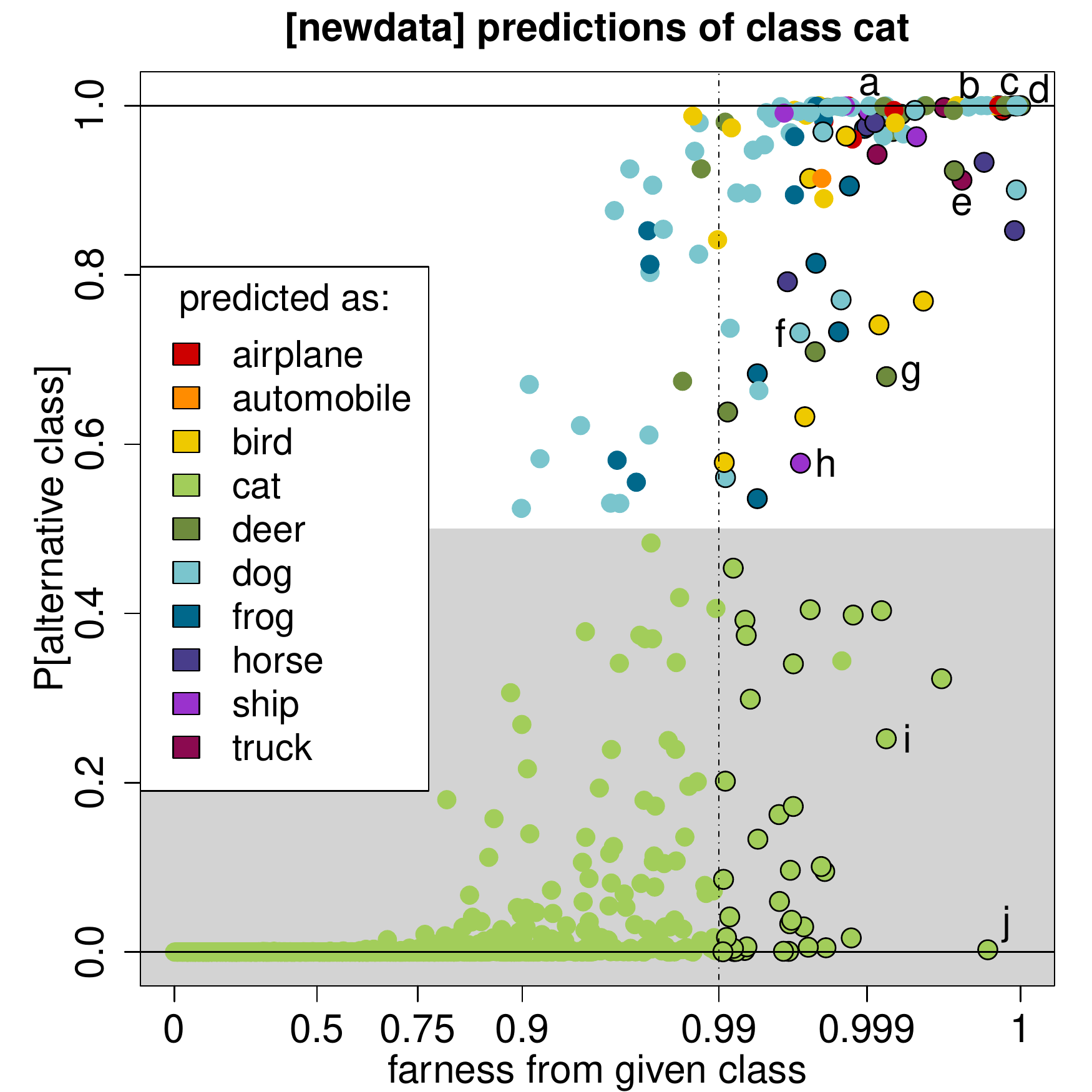}\\
\vspace{0.3cm}
\includegraphics[width = 0.8\textwidth]
  {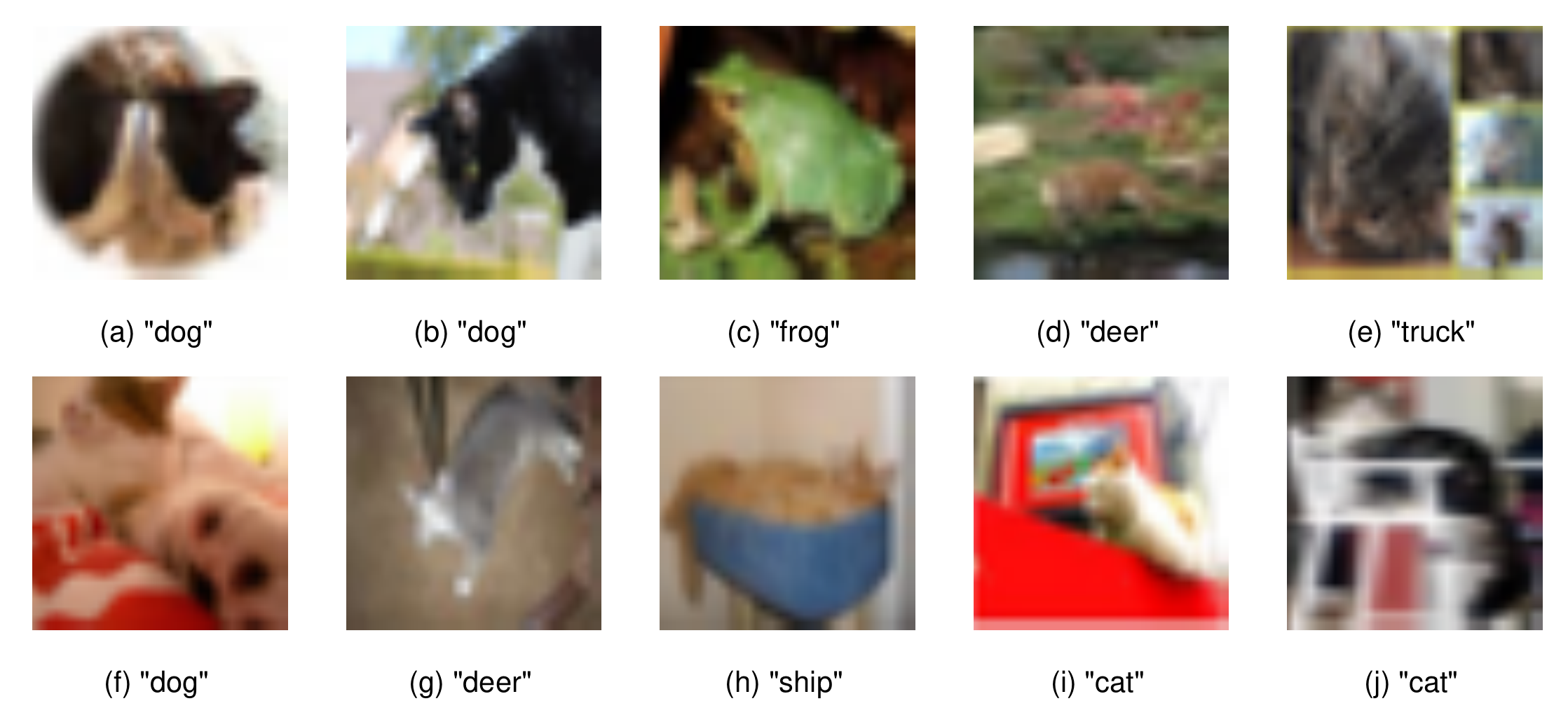}
\caption{Class map of the cat class in the test data,
 with the corresponding images.}
\label{fig:cifar_classmap_cat}
\end{figure}
   
As a second example we consider the class map of 
cat images, shown in 
Figure~\ref{fig:cifar_classmap_cat}. As could be 
expected, there is some confusion with the 
dog class. Not only do cats and dogs often have 
similar features, they also tend to be photographed
with similar backgrounds,
as illustrated in the images \texttt{a}, \texttt{b} 
and \texttt{f}. Point \texttt{c} has both a 
very high farness and the highest PAC.
This is clearly a mislabeled image, which should 
belong to the frog class!
We detected this image because of its
extremely high $\PAC$ and farness, whereas it
would have been harder to find if we only
had a long list of misclassified cases.
Image \texttt{d} is classified as deer, but flagged 
as a farness outlier, suggesting that this image 
does not lie well within any class.
Images \texttt{e} and \texttt{h} are misclassified
as a truck and a ship respectively. 
The first may be explained by the straight lines in 
the image, whereas the second is of a cat in a blue
container. As both of these are farness outliers, 
they are not close to any of the classes, making
them hard to classify.
Finally, images \texttt{i} and \texttt{j} 
are farness outliers but classified correctly.
They contain cats, but the images are dominated
by other objects. 

In both class maps, higher $\PAC$ values 
are associated with higher farness. 
This suggests that in the CIFAR-10 dataset, the
inaccuracy of the neural net classifier is 
caused more by {\it feature noise} (atypical 
images) than by {\it label noise} (randomness 
in the response).

Some other classes are shown in 
section~\ref{suppmat:cifar} of the supplementary
material.

\subsection{Classification trees} \label{sec:rpart}

\begin{figure}
\centering
\includegraphics[width=0.7\columnwidth]
   {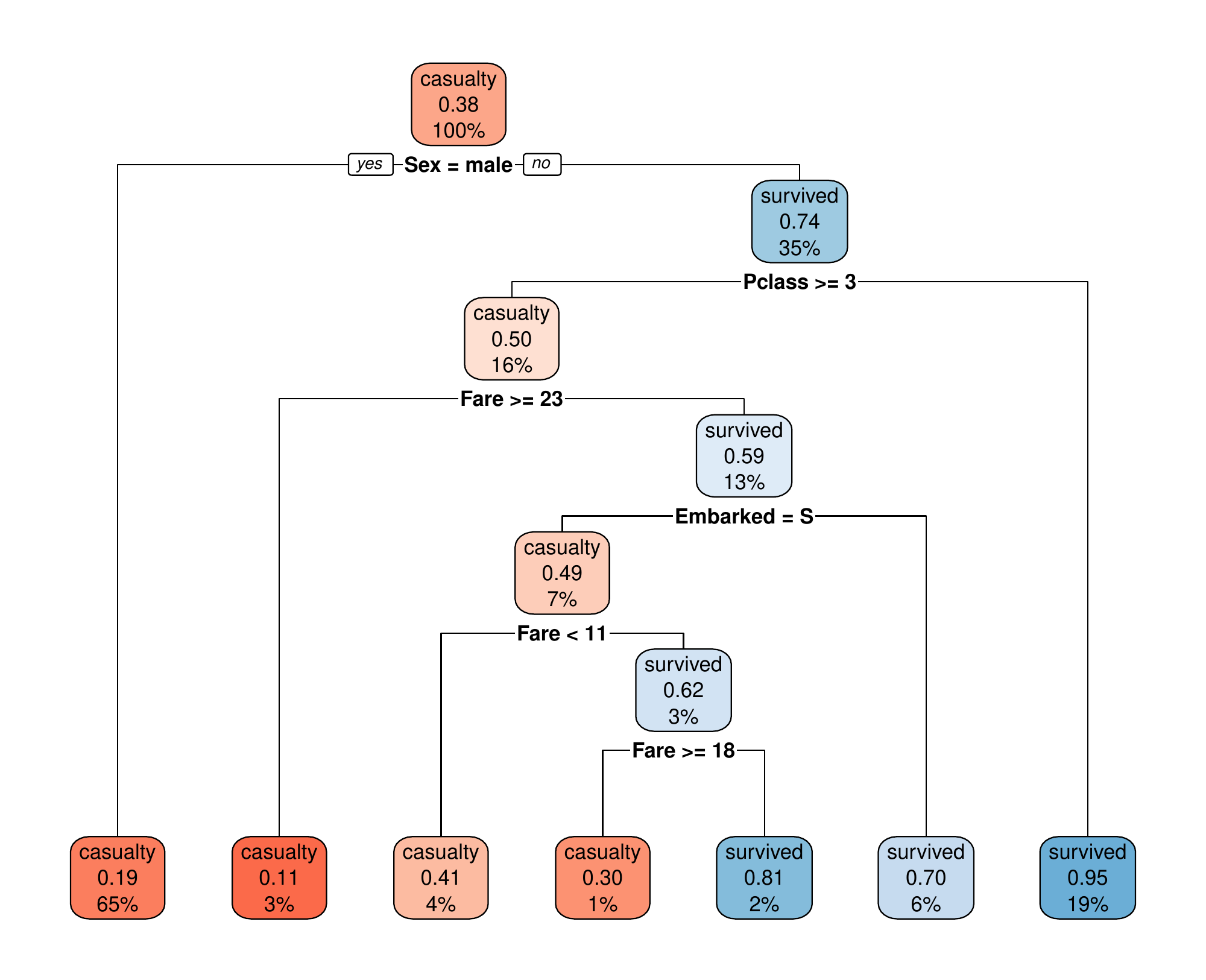}
\caption{CART classification tree trained 
          on the Titanic data.} 
\label{fig:titanic_tree}
\end{figure}

In this section we will visualize the cases in a 
tree-based classification.
Here we use CART \citep{CART}, for which many
implementations exist such as the \textsf{R} 
package \texttt{rpart} \citep{rpart}, but other
tree-based classifiers such as C4.5 \citep{C4.5}
can be visualized as well.
As an illustrative example we analyze the 
Titanic data. 
This dataset is freely available on 
\url{https://www.kaggle.com/c/titanic/data}\,. 
The data contains information on the passengers 
of the RMS Titanic. 
The binary response variable indicates whether
the passenger survived or was a casualty. 
It also contains a mix of nominal, ordinal and 
numerical variables describing several 
characteristics. 
Strong points of CART are its ability to deal with
features of mixed types as well as missing values,
which are abundant in these data.
We train the classification tree 
predicting the survival of the passengers 
from the features \texttt{Pclass}, \texttt{Sex}, 
\texttt{Sibsp}, \texttt{Parch}, \texttt{Fare} and 
\texttt{Embarked}. \texttt{Pclass} is an ordinal 
variable ranging from 1 (first class) 
to 3 (third class), \texttt{Sex} is 
male or female, and \texttt{Fare} is in British
Pounds.  
The variables \texttt{Sibsp} 
and \texttt{Parch} count the number of
siblings+spouses and parents+children aboard. 
\texttt{Embarked} is the port (out of three) at 
which the passenger got on the ship.  
The resulting tree is shown in 
Figure \ref{fig:titanic_tree}, 
drawn with the \texttt{rpart.plot} package 
\citep{rpartplot}.
We see that only 4 out of 6 variables are actually 
used in the model, and that the tree starts with a 
very crude split which predicts all males as 
casualties. This tree has an accuracy of about 82\% 
on the training data.

At the bottom of the tree in 
Figure~\ref{fig:titanic_tree} we see the leaves.
The leftmost leaf contains all males, which make up
65\% of the passengers, with the probability of 
survival being 19\%. So for all cases $i$ that end 
up in this leaf, the posterior probabilities are
\begin{equation*}
  \hp(i,\mbox{survived}) = 0.19 
	\;\;\;\; \mbox{ and } \;\;\;\;
	\hp(i,\mbox{casualty}) = 1 - 0.19 = 0.81\;\;.
\end{equation*}
The classification by the maximum a posteriori
rule~\eqref{eq:MAP} thus assigns all objects in
this leaf to the casualty class, which is listed 
as the top line inside the leaf.
Analogously, the rightmost leaf represents 19\%
of all passengers, its posterior probability of 
survival is 95\%, so all members of this leaf
are predicted as survived.

\begin{figure}
\centering
\includegraphics[width=0.65\columnwidth]
   {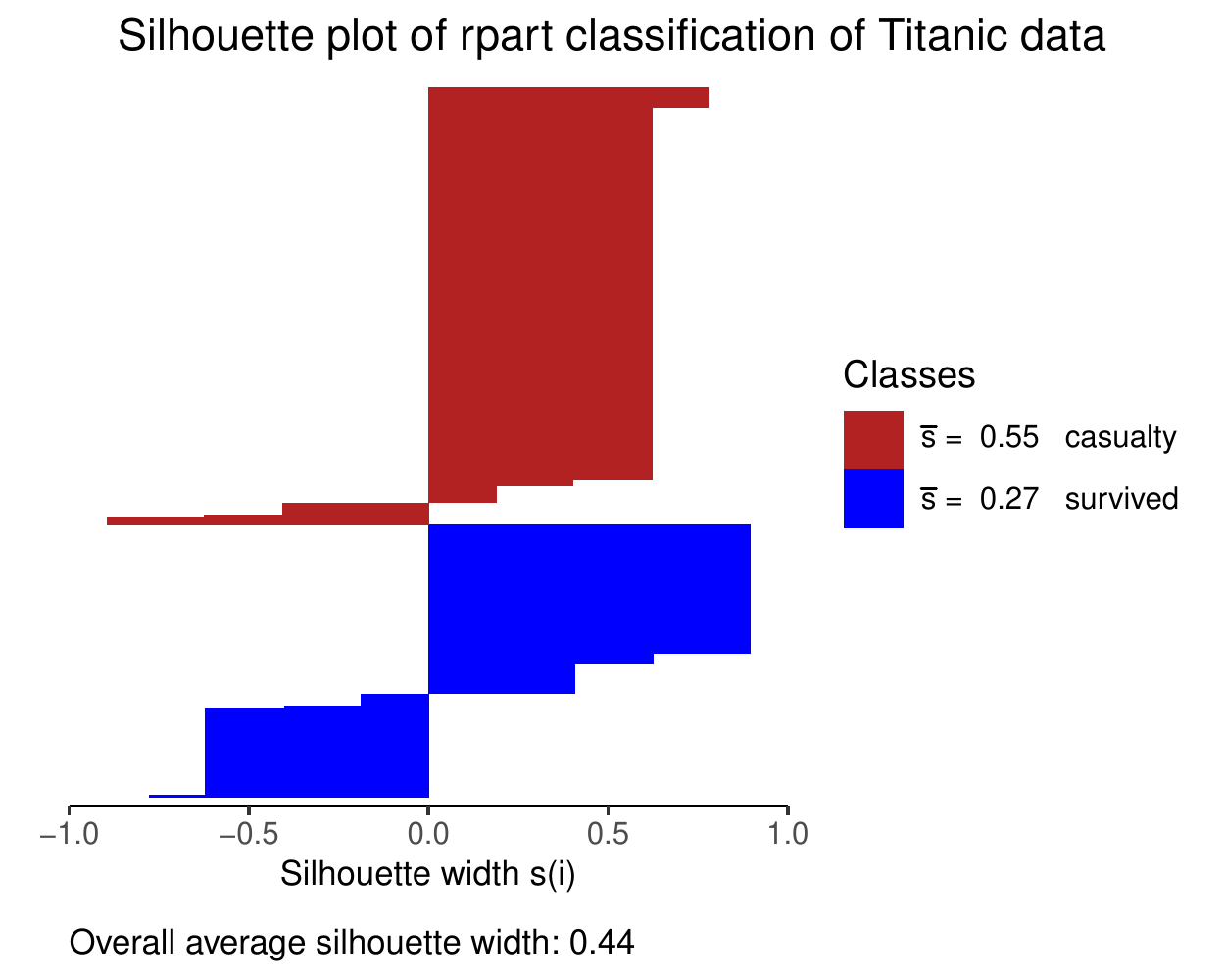}
\caption{Silhouette plot of the CART 
classification on the Titanic training data.}
\label{fig:titanic_sil}
\end{figure}

Now that we know the posterior probabilities
for each case $i$ in the dataset, it is trivial
to compute the probability of the alternative 
class $\PAC(i)$ from~\eqref{eq:PAC}.
Next, \eqref{eq:silclassif} immediately yields 
the silhouette plot, shown in
Figure~\ref{fig:titanic_sil}.
With an average silhouette width of 0.44, we
conclude that this classification of the Titanic 
data is far from perfect, but it may be hard to do
much better given the presumably chaotic decision
making at the time of the disaster.
By comparing the average silhouette widths, we
see that the class of survived passengers 
(in blue) is hardest to predict.
But the correct predictions in this class are made 
with a relatively high conviction 
($s(i) \approx 1$). This is in contrast to 
the predictions of the casualty class which 
contains fewer misclassified cases, but the correct 
classifications for this class are made with only 
moderate conviction.

\begin{figure}[!ht]
\centering
\includegraphics[width = 0.54\textwidth]
  {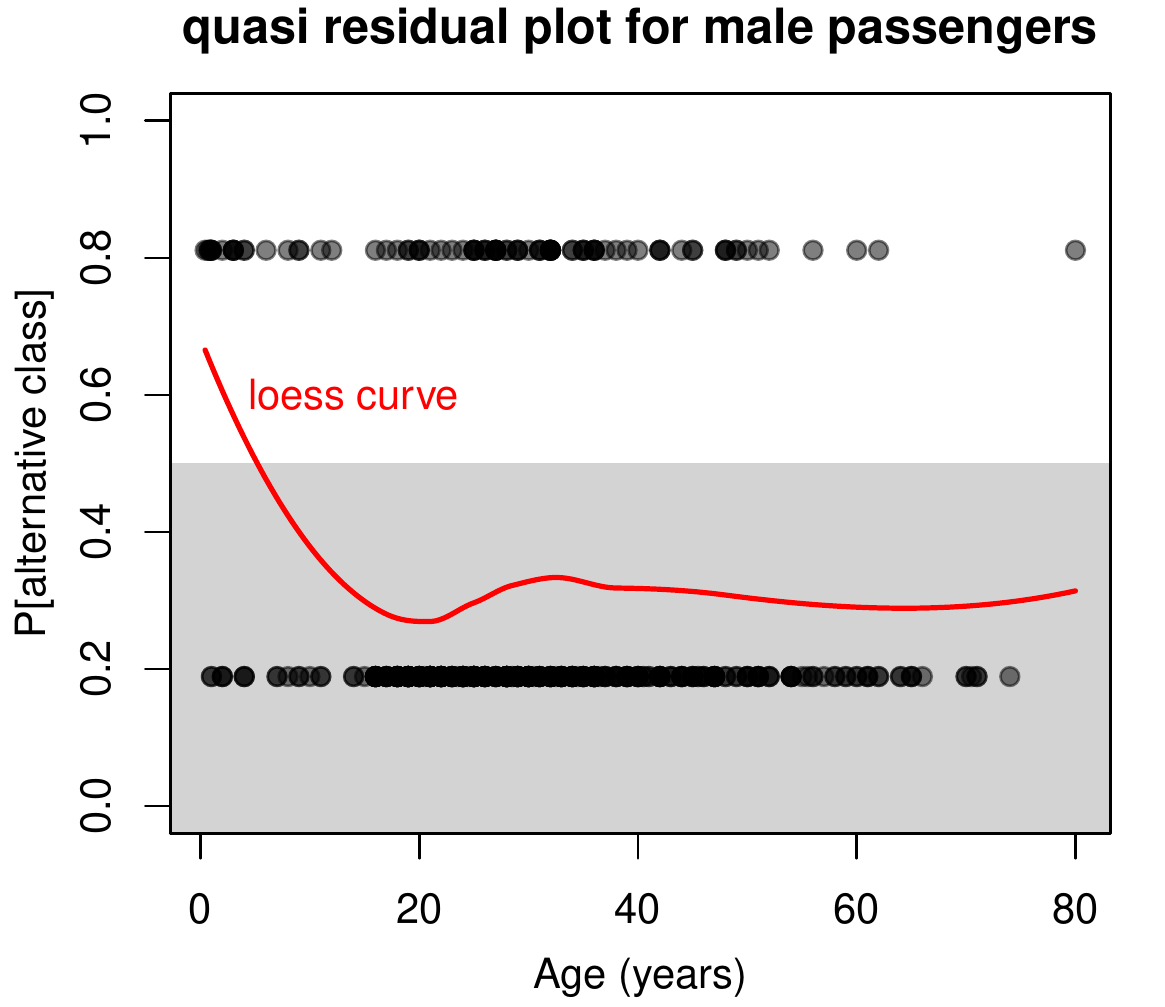}
\caption{Quasi residual plot of male passengers
         versus their age, with loess curve in red.}
\label{fig:titanic_qrplot}
\end{figure}

Figure \ref{fig:titanic_qrplot} shows an interesting 
quasi residual plot, of PAC versus age for the males 
in the data. 
The PAC only takes two values in this subset of the
data, corresponding to the leftmost leaf in 
Figure~\ref{fig:titanic_tree}.
As a visual aid the loess curve~\citep{loess} was
superimposed, using the \texttt{loess} function
in base \textsf{R} with default settings.
This indicates that the PAC values (here linked to
survival) for very young males are 
substantially higher than for older males. 
The graph thus uncovered a phenomenon in a
subset of the data. 

Note that each leaf of a classification tree such 
as Figure~\ref{fig:titanic_tree} corresponds to a 
number of splits in the features, so its domain 
in feature space can be called `rectangular'. 
The region that is assigned the same prediction 
is thus a union of such `rectangles', which does 
not need to be connected.
Therefore, a tree-based classifier takes a 
completely different view of the data than,
for instance, linear discriminant analysis.
In order to construct class maps, we need a 
measure of farness in feature space which is in 
line with how the classifier looks at the data. 
Therefore, it is natural to take a distance measure 
that is additive in the features. Moreover, the 
distance measure needs to be able to handle features
of mixed types, as well as missing values. 
For these reasons we elect to use the 
daisy function introduced by \cite{FGID},
which is a dissimilarity version of the 
similarity coefficient of \cite{gower1971general} 
for nominal, asymmetric binary, and numerical 
variables, extended to encompass ordinal variables. 
It is implemented as the function \texttt{daisy} in 
the \textsf{R} package \texttt{cluster}
\citep{cluster_package}.
 
Simply applying \texttt{daisy} to the set of 
features would ignore an important property of
tree-based classifiers, which is that they do not
consider all features equally important.
To take this aspect into account, we use a 
weighted daisy dissimilarity where the weights 
are equal to each variable's importance.
The variable importance is a standard output 
of \texttt{rpart}, computed as described in 
\citep{CART}.
The weighted daisy computation provides us with a 
dissimilarity $d(i,j)$ between all cases $i$ and $j$.
When computing the farness of a case $i$ to a 
class $g$, we need to take into account the local 
nature of classification trees.
To this end we use a nearest-neighbors type 
approach, described in detail in 
section~\ref{suppmat:daisyfarness}.

\begin{figure}
\includegraphics[width=1.00\columnwidth]
  {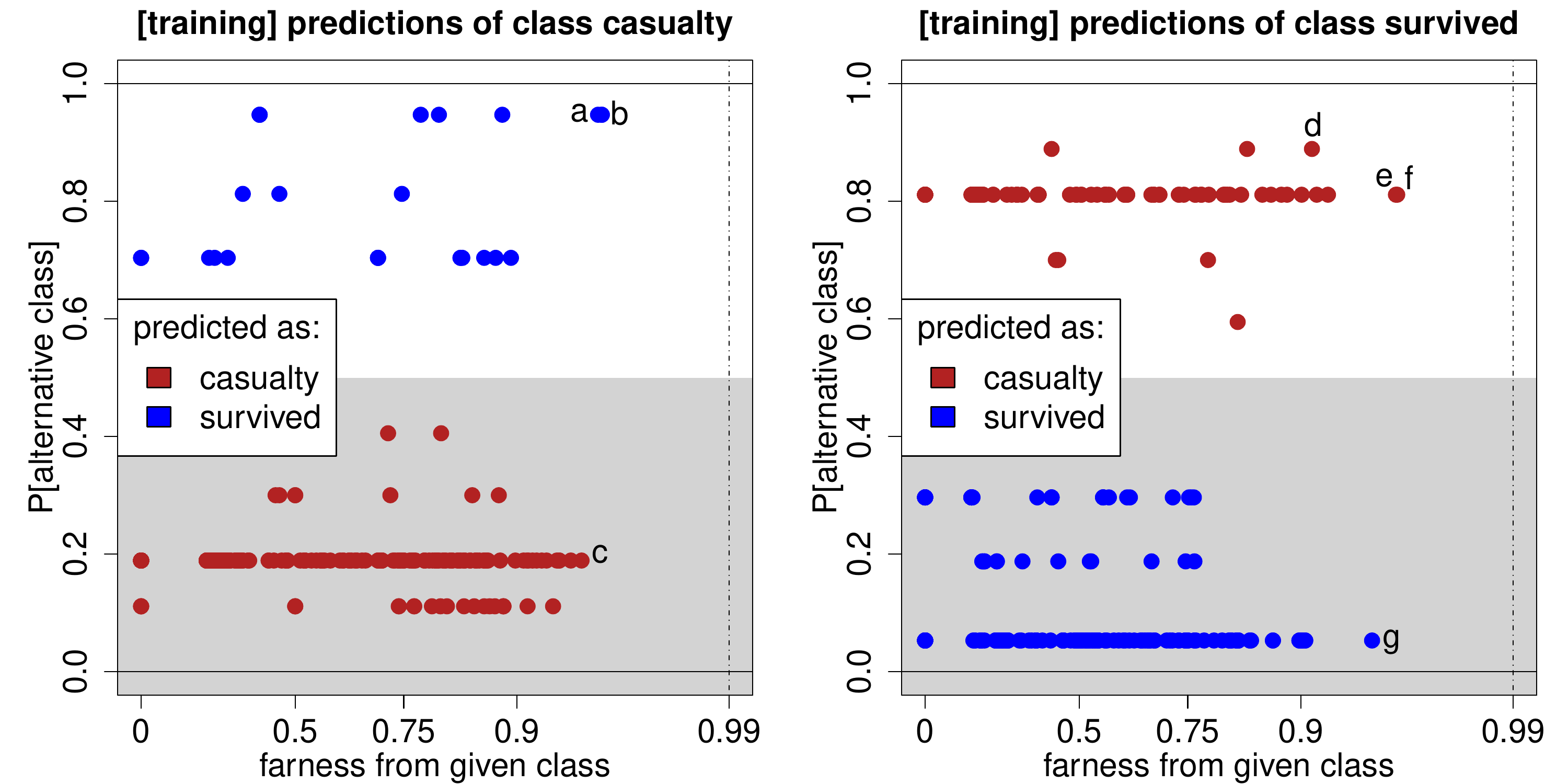}	
\caption{Titanic training data: class maps of  
  casualties (left) and survivors (right).}
\label{fig:titanic_classmaps}
\end{figure}

Figure \ref{fig:titanic_classmaps} shows the 
class maps of the Titanic training data, in 
which we note a few extreme points. 
Cases \texttt{a} and \texttt{b} have 
the highest PAC in the casualty class,
combined with a relatively high farness. 
These are a woman and female child traveling in 
first class, for which the classifier predicted
survival. The elevated farness is due to some
unusual characteristics for the casualty class, 
such as a high fare for the child, the gender 
and traveling class of both subjects, and an 
uncommon port of embarkation for the woman. 
Passenger \texttt{c} has a low PAC but a rather
high farness. This is a male passenger who was 
correctly predicted as a casualty. His high
farness is caused by paying a huge fare (in the
top 1\%), traveling in first class, and having as
many as 4 children+parents traveling with him.

The class map of the survived passengers is shown 
in the right hand panel of 
Figure \ref{fig:titanic_classmaps}.
Case \texttt{d} is a woman who traveled in third 
class and is misclassified as a casualty.
Her relatively high farness is caused by the 
fact that she was traveling with as many as
5 parents+children.  
Passengers \texttt{e} and \texttt{f} are two males, 
and thus predicted as casualties with high PAC. 
Their substantial farness is explained by having
paid a high fare. 
Point \texttt{g} is a woman traveling first class, 
correctly predicted as survived, with much
conviction since her PAC is close to zero.
She paid the highest fare of all passengers
in the training data, causing her relatively 
high farness.

In Figure \ref{fig:titanic_classmaps}
we note that high $\PAC$ values occur over
the whole farness range, telling us that for
the Titanic data the misclassifications were
mainly caused by label noise, i.e. much
randomness in the survival label.
Subsection~\ref{suppmat:titanic} shows the
corresponding class maps for the Titanic test 
data, where again some individuals stand out.

\subsection{Random forests} \label{sec:RF}

Random forests were introduced by \cite{Breiman:RF} 
and are based on an ensemble of decision trees.
The idea is to train many different
classification trees for the same task.
In order to generate sufficiently diverse trees, 
two techniques are exploited. The first is bagging,
which means that only a subsample of the observations 
is used when training a single tree. The second is the 
random sampling of potential variables at each split. 
This forces the various trees to use a wide variety 
of variables.
To classify case $i$, we let it go down all of the
trees in the forest. Its posterior probability
$\hp(i,g)$ in class $g$ is then the number of trees
that assigned it to class $g$, divided by the total 
number of trees.
These posterior probabilities clearly add up to 1.
Applying the maximum a posteriori 
rule~\eqref{eq:MAP}, we then assign case $i$ to the
class with highest $\hp(i,g)$. 
Random forests often perform well in real 
world classification problems. Here we use the 
implementation in the \texttt{R} package
\texttt{randomForest} by \cite{R:randomForest}.

As an illustration we analyze the emotion dataset 
of \cite{mohammad-bravo-marquez-2017-wassa}.
It contains a training set of 3613 tweets and a 
test set of 3142 tweets, which have been labeled
with the four classes anger, fear, joy, and 
sadness. 
The goal is to train a random forest to predict 
the emotion of a tweet. 
We preprocessed the data by removing
word contractions and elongations using the 
\textsf{R} package \texttt{textclean} 
\citep{R:textclean}.
We also replaced the emojis with unique words.
Finally, we used the \textsf{R} package 
\texttt{text2vec} \citep{R:text2vec} to convert
these texts into numerical data based on n-grams 
of at most length~3. This procedure leaves us 
with a $3613 \times 2705$ training data matrix. 
The class anger has 857 tweets, fear has 1147,
joy has 823, and sadness has 786. 
We then ran \texttt{randomForest()} 
with its default options.
The trained forest achieves an accuracy of 
97.6\% on the training data, and 
80.8\% on the test data. 
Since each case $i$ gets posterior 
probabilities $\hp(i,g)$ for each class $g$,
we can easily compute the probability of the
alternative class $\PAC(i)$ 
from~\eqref{eq:altclass} and \eqref{eq:PAC}.
The silhouette plot for the training data 
was shown in Figure~\ref{fig:silh1}. 

\begin{figure}[!ht]
\centering
\includegraphics[width = 0.85\textwidth]
  {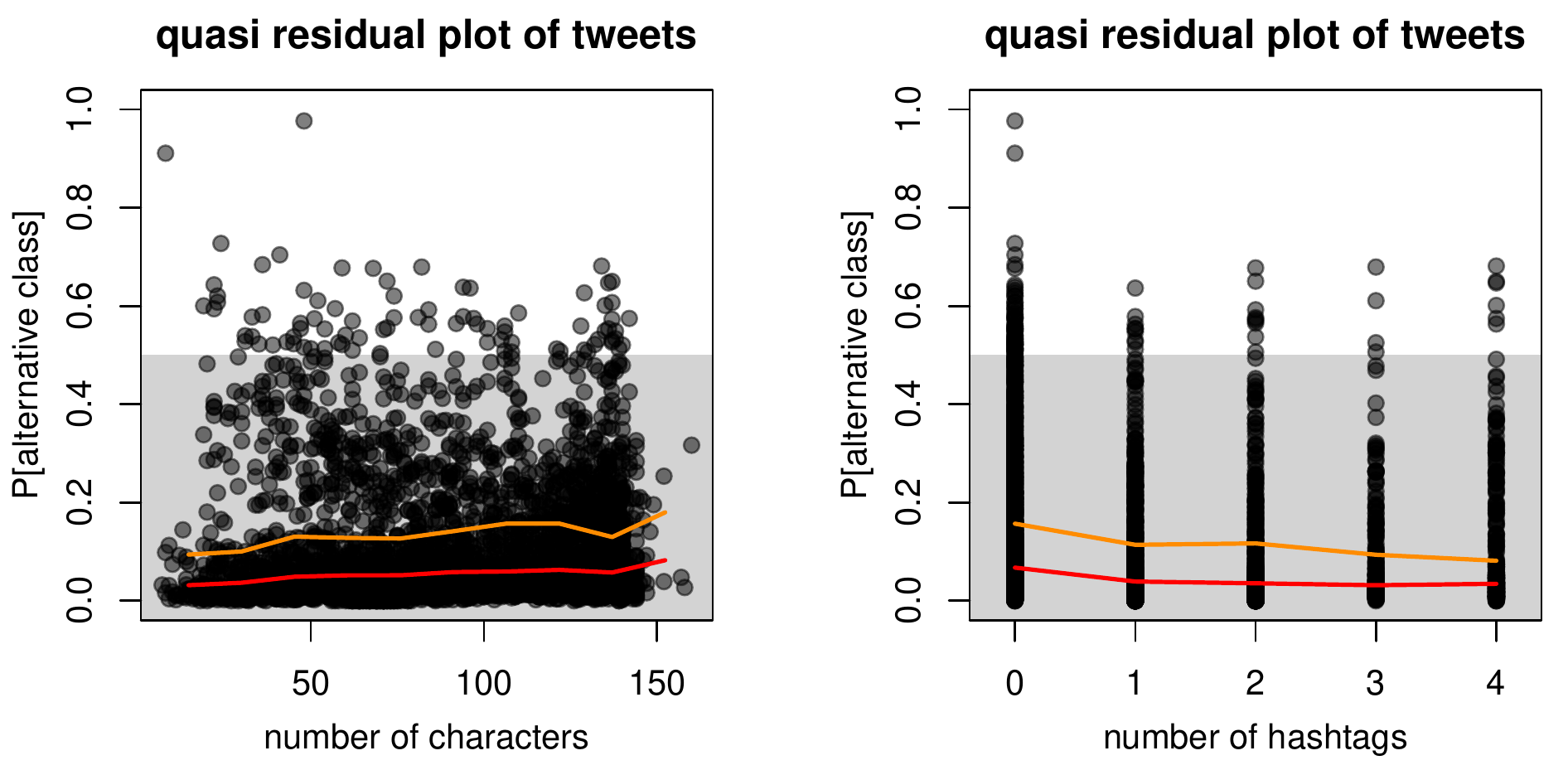}
\caption{Quasi residual plots, with their 
         medians (red) and 75th
				 percentiles (orange).}
\label{fig:emotion_qrp}
\end{figure}				

The left panel of Figure~\ref{fig:emotion_qrp}
is the quasi residual plot of PAC versus the
number of characters in each tweet, with the
median (in red) and 75th percentile (orange)
on 10 equispaced intervals. 
As most tweets are classified correctly with 
low PAC values, these trend lines are near
the bottom of the plot. But we still see an
upward trend, which indicates that longer tweets 
were somewhat harder to classify, perhaps due
to containing words linked with more than one 
class. 
The quasi residual plot in the right hand panel
is versus the number of hashtags (0, 1, 2, 3, 
and 4+). This time we see a downward trend,
suggesting that tweets with more hashtags are 
typically easier to classify.
Both plots revealed an effect that we
could not have predicted beforehand.
	
\begin{figure}
\centering
\includegraphics[width=0.98\columnwidth]
  {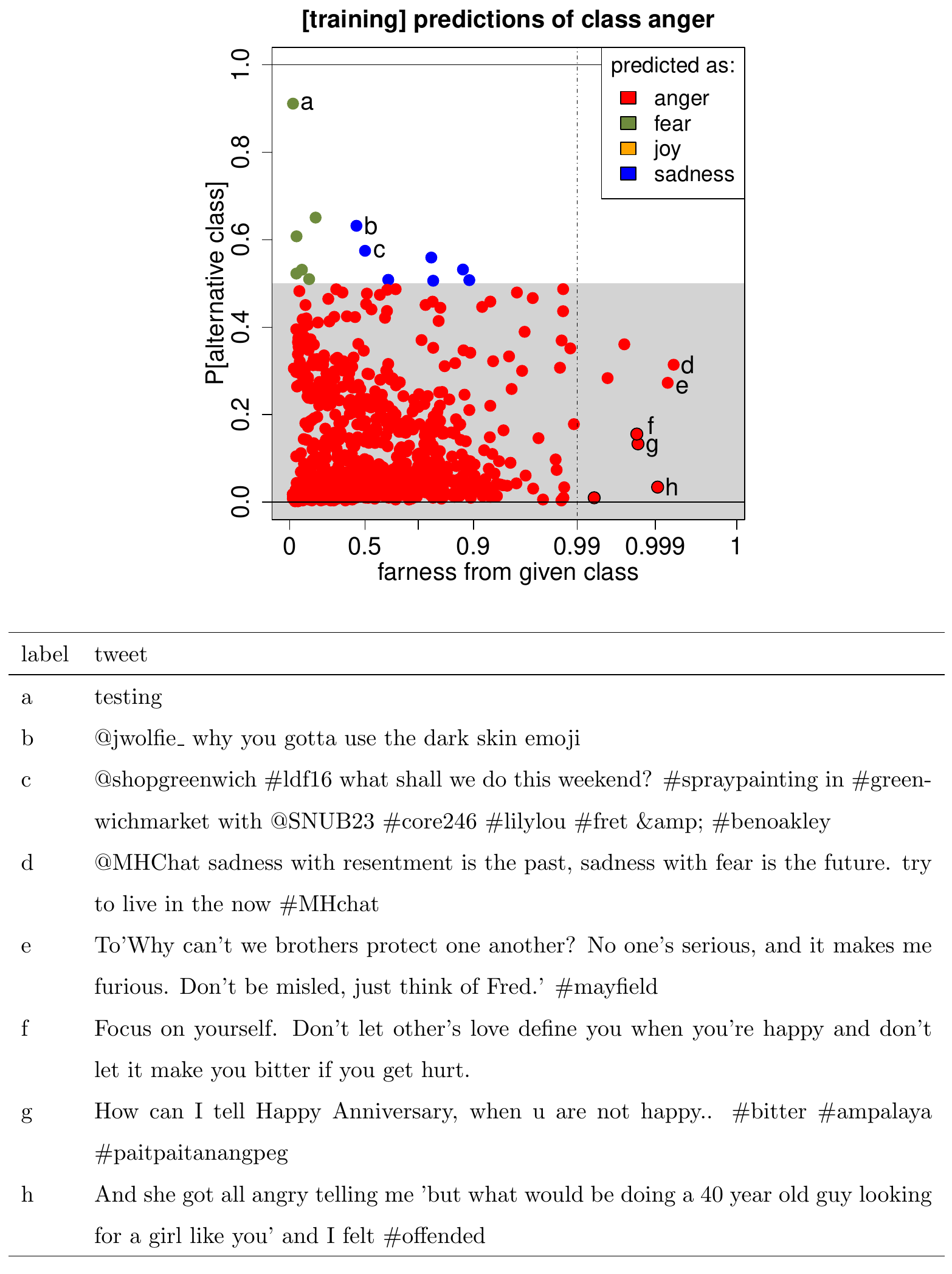}
\caption{Class map of the anger class, with
  the corresponding tweets.}
\label{fig:emotion_classmap_anger}
\end{figure}
			
We now turn to the class maps of the emotion data.
The farness is computed along the same
lines as in the previous section, starting from a
weighted daisy dissimilarity between cases.
The weights are again given by each variable's
importance, corresponding to the total decrease
of the Gini index by splitting on the variable,
and averaged over all trees in the forest. 
This is a standard output of the function 
\texttt{randomForest()}.
Deriving the farness of case $i$ from class $g$
is also done in the same way, described in 
section~\ref{suppmat:daisyfarness}
of the Supplementary Material.

Figure~\ref{fig:emotion_classmap_anger} shows the 
class map of the anger class. 
Only a few points aren't red, meaning they are 
assigned to a different class.
We marked some points that jump out, and the 
corresponding tweets are listed below the class map.
Point \texttt{a} has the highest PAC in this class. 
It corresponds to the uninformative tweet `testing' 
which does not contain any word in the constructed
vocabulary, so it gets assigned to the fear class
simply because that class has the most members. 
The class map drew our attention to
this atypical tweet.
Tweets \texttt{b} and \texttt{c} are assigned to
sadness. The first might indeed be sad, but the
context is lacking.
The second does not seem to carry a clear emotion.
We also look at some tweets that are classified 
correctly, but lie far from their given class. 
Tweet \texttt{d} seems more sad than angry, and
contains a lot of words which would not 
immediately be associated with anger. 
Tweet \texttt{e} is in fact part of a song lyric.
Finally, \texttt{f}, \texttt{g} and \texttt{h} 
have a black border which indicates that
their farness to {\it all} classes is above
the cutoff value, so they do not lie
well within any class. Indeed, they refer to
emotions belonging to none of the classes, such 
as bitterness and feeling offended.

\begin{figure}
\centering
\includegraphics[width=0.98\columnwidth]
  {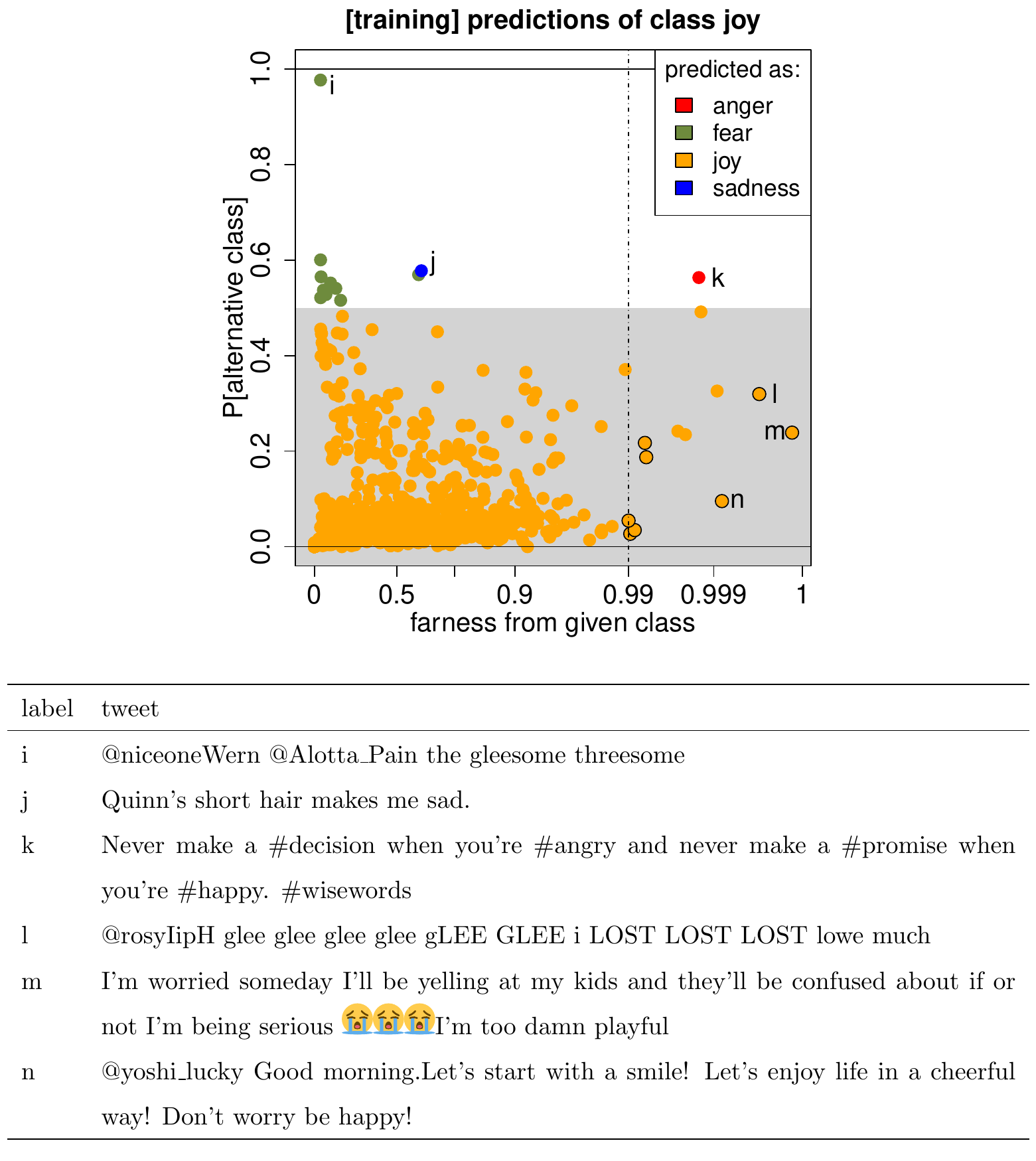}
\caption{Class map of the joy class, with
  the corresponding tweets.}
\label{fig:emotion_classmap_joy}
\end{figure}

The class map of the joy class is shown in 
Figure~\ref{fig:emotion_classmap_joy}.
There are only a few misclassified points, 
as well as a handful of farness outliers on the 
bottom right. 
Tweet \texttt{i} contains the word gleesome,
which does suggest joy, but this word only occurs 
once in the dataset so it is not in the 
constructed vocabulary, leading \texttt{i} 
to be assigned to the largest class (fear).
Tweet \texttt{j} appears to be mislabeled, as 
it suggests sadness instead of joy.
Tweet \texttt{k} is classified as angry, but 
not with high conviction since its $\PAC(i)$
is only slightly above 0.5\,. It also has a 
rather high farness, indicating that it doesn't 
lie well within the joy class. The text is
a proverb about two emotions.
Tweets \texttt{l} to \texttt{n} have a 
black border indicating farness outliers, 
which suggests that they do not lie 
well within any of the classes.
Tweet \texttt{l} is indeed strange with many
repetitions, and \texttt{m} mixes emotions
so it is hard to give it a single label.
Tweet \texttt{n} is definitely in the joy 
class, but contains an unusual number of 
joy-related words compared to other tweets 
of this class.

In both class maps the $\PAC > 0.5$ 
values are not associated with high farness.
This suggests that the misclassifications 
are mainly driven by label noise, caused by 
the difficulty of labeling the emotion of
some tweets.
The class maps of the remaining emotions fear
and sadness are discussed in 
section~\ref{suppmat:emotion} of the
supplementary material.

\section{Conclusions} \label{sec:conclusions}

The proposed visualizations focus
on the cases in a classification.
The examples illustrated the
benefits of this approach.
The new silhouette plot describes the strength 
of each object's classification, grouped by 
class. For instance, we noticed that
images of mechanical objects were typically
classified more reliably than images of 
animals.
Quasi residual plots yielded other insights, 
such as trends in subsets of the data like 
the effect of age for  
male passengers on the Titanic. 
They also revealed factors affecting the 
classification accuracy, such as the length 
of tweets and their number of hashtags.
The class map provides additional information,
as it can tell us which cases lie between 
classes, which cases are far from their given 
class, and some cases may be far from all 
classes. The class map allowed us to
distinguish between feature noise and
label noise in the examples:
in the image data the misclassifications
were mainly driven by atypical images,
whereas in the other examples the dominant
effect was some randomness in the response 
(the labels), such as survival in the 
Titanic data.
The displays also drew our attention 
to atypical cases
that were inspected in more detail,
providing further insights in the data.

The visual displays in this paper
were produced with the R package 
\texttt{classmap}\linebreak
\citep{classmap} on CRAN. Its vignettes 
\texttt{Neural$\_$net$\_$examples},\linebreak
\texttt{Rpart$\_$examples} and 
\texttt{Random$\_$forest$\_$examples}
correspond to the classifiers in section
\ref{sec:classifiers}.
Note that all three visualizations 
make use of the posterior 
probabilities that a case belongs to the 
available classes (labels). 
Since most classifiers provide such
probabilities, the graphical displays can be
employed not only with neural nets and tree-based
classifiers, but also with other methods
such as discriminant analysis, k-nearest 
neighbors, and support vector machines. The
displays for these methods are also available 
in the {\it classmap} package.

\vskip0.4cm 


\noindent{\bf Software availability.}
 An \textsf{R} script reproducing the examples
 in this paper can be downloaded from
\url{https://wis.kuleuven.be/statdatascience/robust/software }.

\vskip0.4cm 

\noindent \textbf{Acknowledgment.} 
This research	was funded by 
projects of Internal Funds KU Leuven.	
The reviewers made helpful
comments improving the presentation.


\clearpage
\pagenumbering{arabic}
\appendix
\numberwithin{equation}{section} 
\section{Supplementary Material} \label{sec:A}
\renewcommand{\theequation}
   {\thesection.\arabic{equation}}
	
\subsection{Fitting a distribution to distances}
\label{suppmat:transfo}

The definition of farness~\eqref{eq:farness}
requires an estimated cumulative distribution
function of the distance $D(\by,g)$ where $\by$ 
is a random object generated from class $g$\,.
The available data are the $D(i,g_i)$ of each 
object $i$ to its class label $g_i$\,.
In view of possible heteroskedasticity between
classes, we start by normalizing per class.
For a given class $g$ we divide all the
$D(i,g)$ where $i$ is a member of class $g$ by 
$\median\{D(j,g)\,;\,j 
 \mbox{ belongs to class } g\}$.
The resulting distances are more homoskedastic,
and we pool them to obtain distances
$d_i$ for $i=1,\ldots,n$.
The empirical distribution of the $d_i$ is
typically right-skewed.

In order to account for skewness, we apply
the function \texttt{transfo} of the 
\textsf{R}-package \texttt{cellWise} 
\citep{cellWise} with default options.
This function first standardizes the $d_i$ to
\begin{equation*}
  x_i = \frac{d_i - \mbox{Med}}{\mbox{Mad}}
\end{equation*}
where $\mbox{Med}=\median_{j=1}^n d_i$
and $\mbox{Mad}$ is the median absolute 
deviation given by 
$\mbox{Mad}=1.4826 
 \median_{j=1}^n |d_i-\mbox{Med}|$
as implemented in the standard function 
\texttt{mad()} in \textsf{R}.
Next, \texttt{transfo} carries out the
Yeo-Johnson transform given by
\begin{equation} \label{eq:YJ}
\YJl(x) = 
\begin{cases}
((1+x)^{\lambda} - 1) / \lambda &\mbox{ if } \lambda 
    \neq 0 \mbox{ and } x \geqslant 0\\
\log(1+x) &\mbox{ if } \lambda = 0 \mbox{ and } 
    x \geqslant 0\\
-((1 - x)^{2 - \lambda} - 1)/(2 - \lambda) &\mbox{ if }
    \lambda \neq 2 \mbox{ and } x < 0\\
-\log(1 - x) &\mbox{ if } \lambda = 2 \mbox{ and } x < 0
\end{cases}
\end{equation}
which aims to bring the distribution close
to a normal distribution.
The transformation $\YJl$
is characterized by a parameter $\lambda$
that has to be estimated from the data. 
This estimation is typically done by maximum 
likelihood, but the default in \texttt{transfo} 
is to apply the weighted maximum likelihood 
estimator of \cite{TVCN} which is less 
sensitive to outliers.
The resulting $\YJl(x_i)$ are in turn standardized
by their own $\mbox{Med}$ and $\mbox{Mad}$, 
yielding $z_i$ whose distribution is 
approximately standard normal. 
The estimated cdf of the distances $d_i$ is then 
given by $\hat{F}(d_i) := \Phi(z_i)$ where $\Phi$ 
is the standard normal cdf.

\subsection{More on the CIFAR-10 data}
\label{suppmat:cifar}

Here we show some visualizations of classes
in the CIFAR-10 data that were not in the main 
text.\\

\begin{figure}[!ht]
\vspace{0.3cm}
\centering
\includegraphics[width = 0.6\textwidth]
  {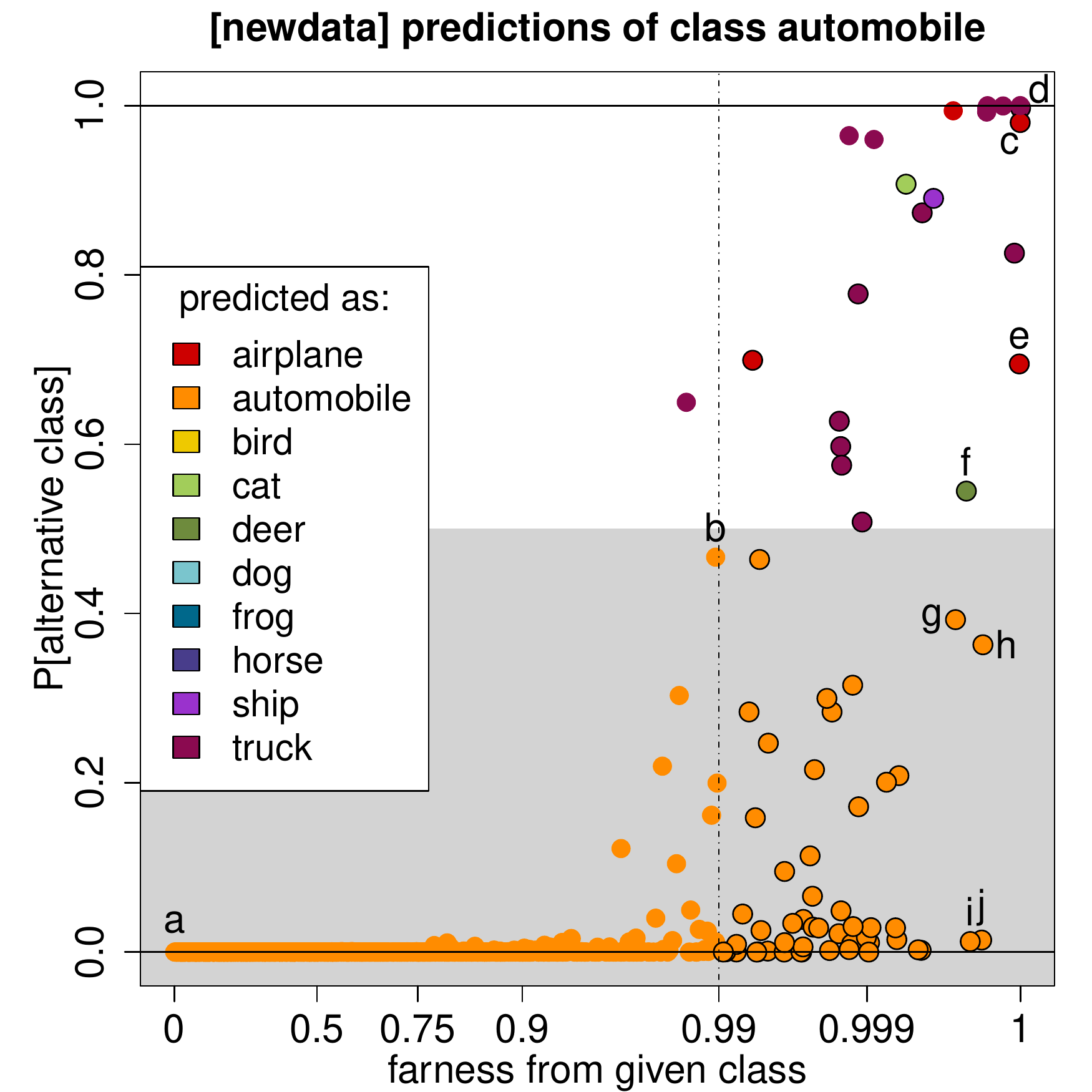}\\
\vspace{0.3cm}
\includegraphics[width = 0.8\textwidth]
  {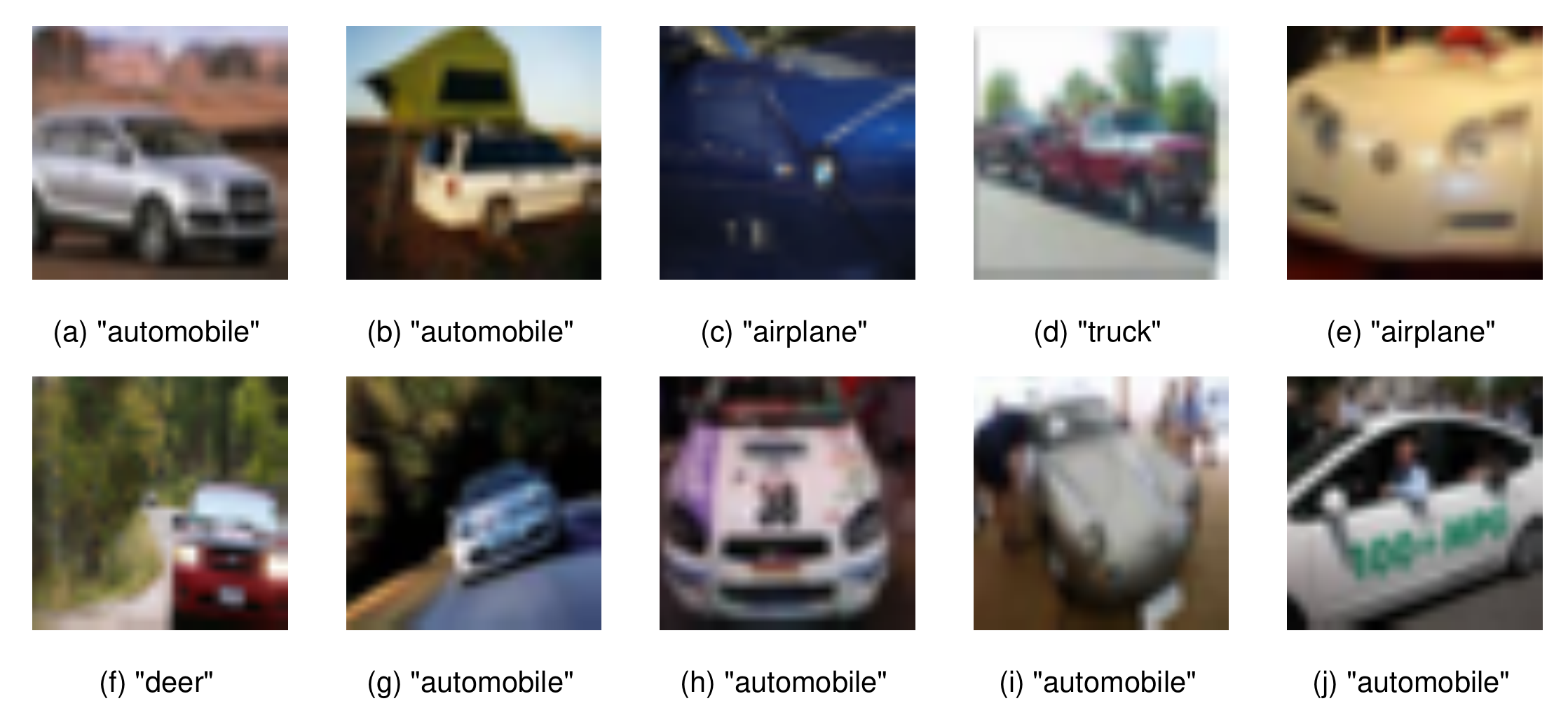}
\caption{Class map of the automobile class, 
  with the corresponding images.}
\label{fig:cifar_classmap_automobile}
\end{figure}

\begin{figure}[!ht]
\vspace{0.3cm}
\centering
\includegraphics[width = 0.6\textwidth]
  {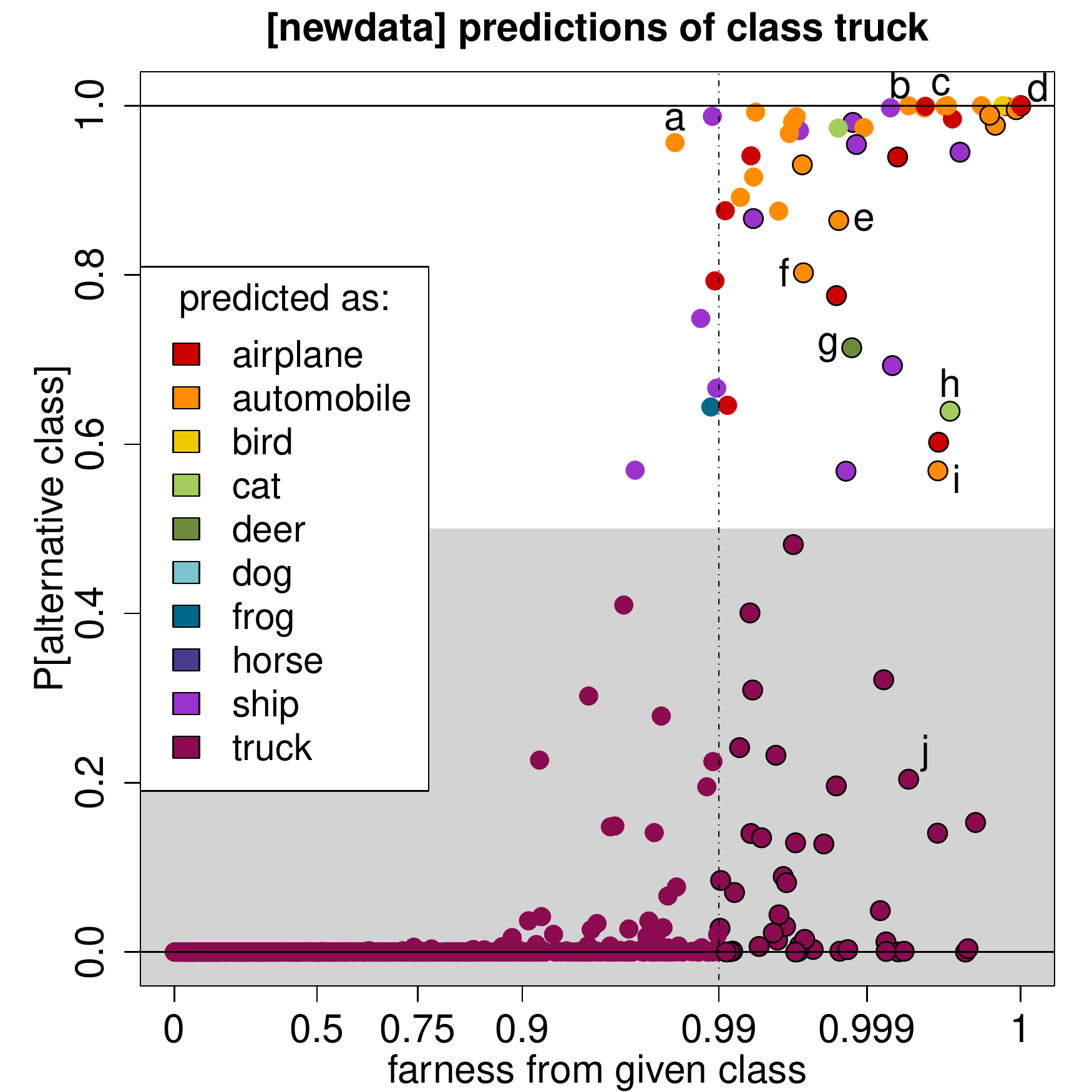}\\
\vspace{0.3cm}
\includegraphics[width = 0.8\textwidth]
  {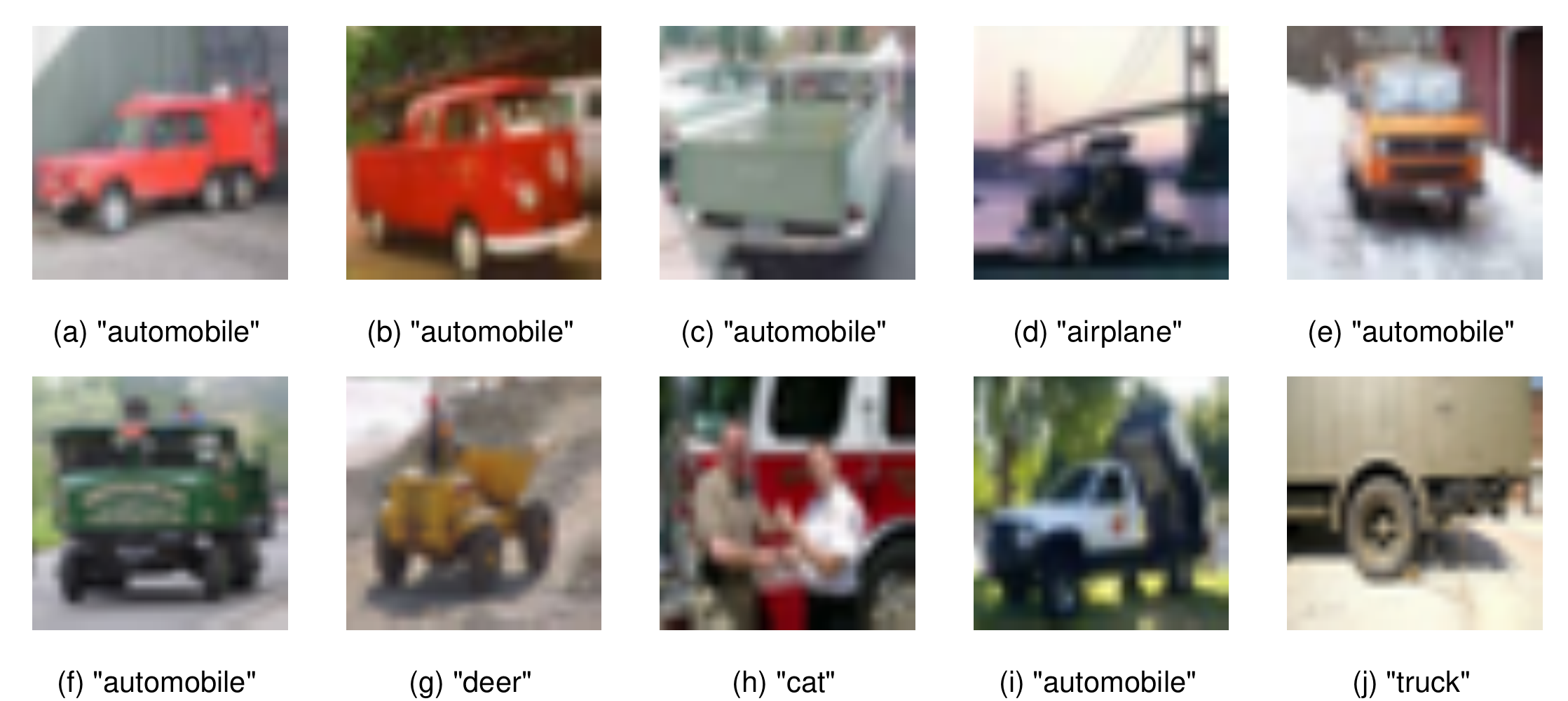}
\caption{Class map of the truck class,
   with the corresponding images.}
\label{fig:cifar_classmap_truck}
\end{figure}

\newpage
Note that (d) in 
Figure~\ref{fig:cifar_classmap_automobile}
and (b) and (c) in
Figure~\ref{fig:cifar_classmap_truck}
look like pickup trucks, which are in a sense
intermediate between automobiles and 
trucks, in spite of the fact that the original 
data description in 
\url{https://www.cs.toronto.edu/~kriz/cifar.html}
aimed to avoid pickup trucks for that reason.\\

\clearpage

\begin{figure}[!ht]
\vspace{2.3cm}
\centering
\includegraphics[width = 0.6\textwidth]
  {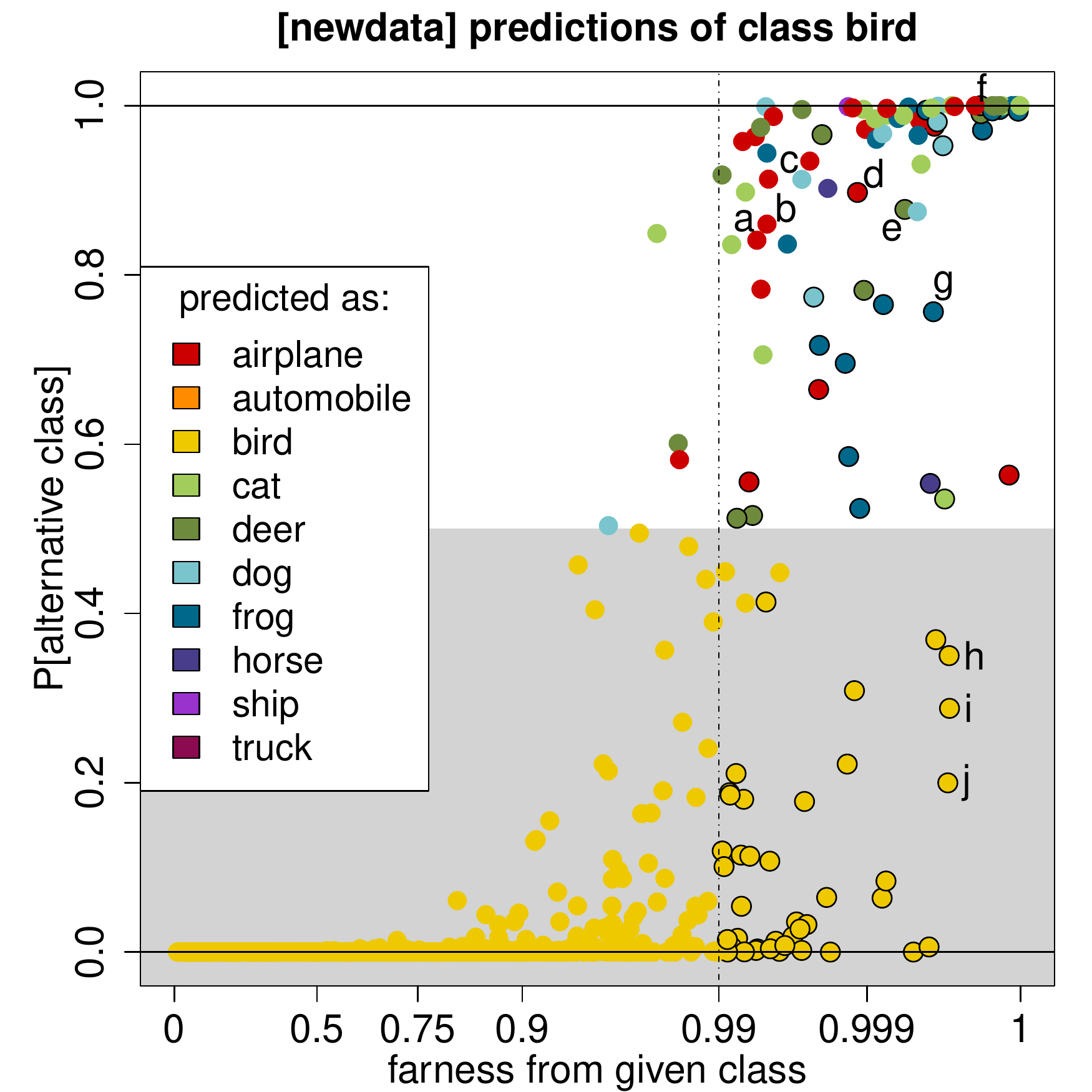}\\
\vspace{0.3cm}
\includegraphics[width = 0.8\textwidth]
  {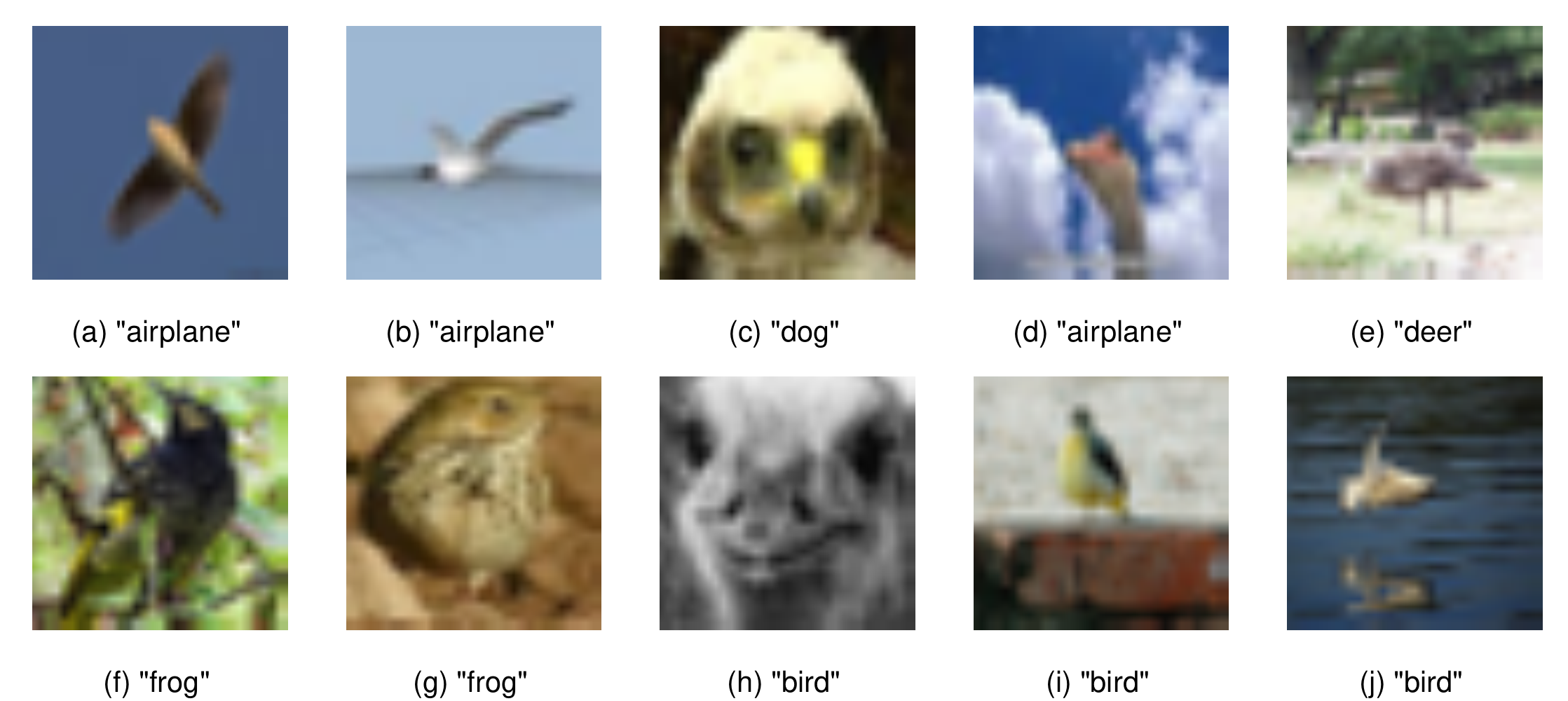}
\caption{Class map of the bird class,
   with the corresponding images.}
\label{fig:cifar_classmap_bird}
\end{figure}
\clearpage

\begin{figure}[!ht]
\vspace{2.3cm}
\centering
\includegraphics[width = 0.6\textwidth]
  {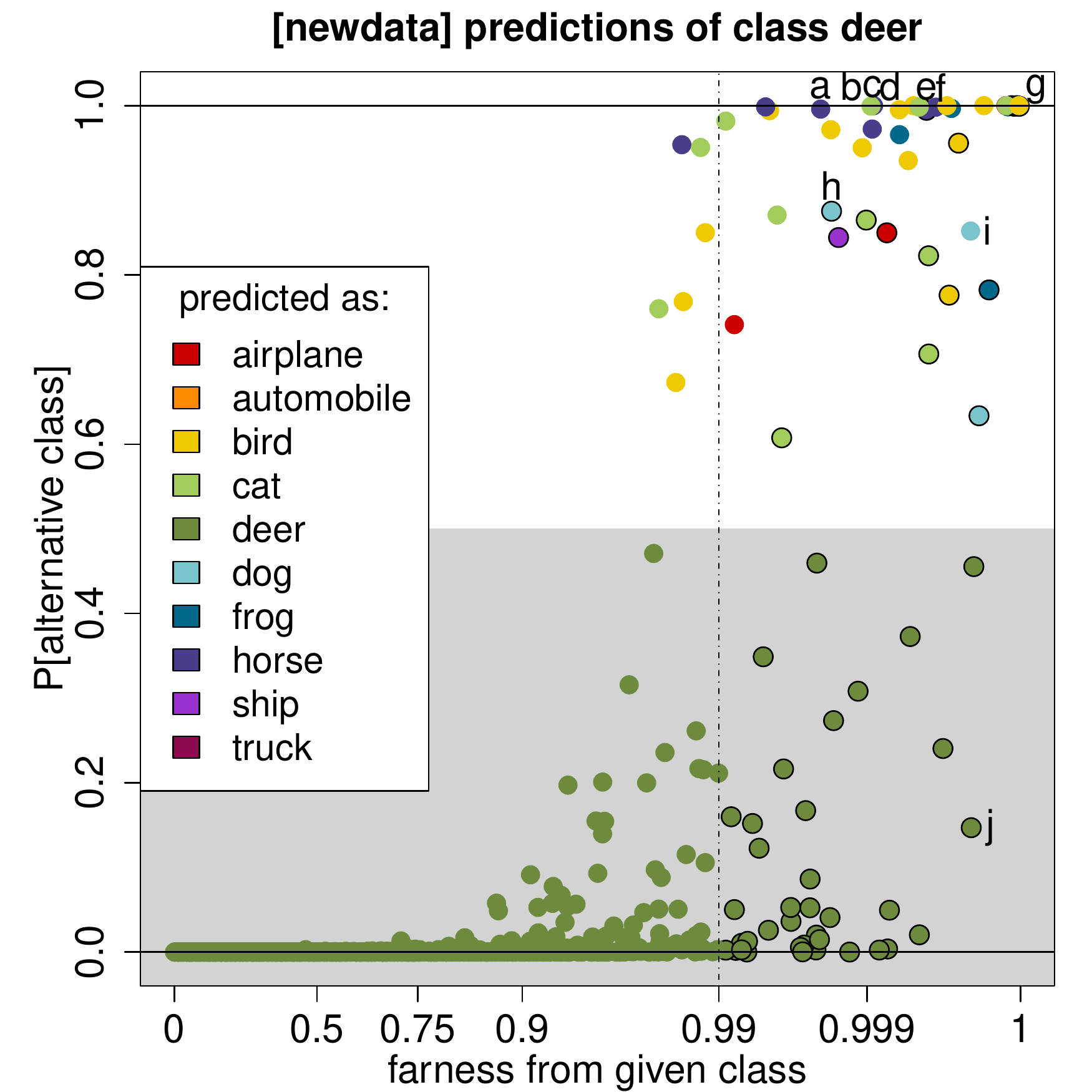}\\
\vspace{0.3cm}
\includegraphics[width = 0.8\textwidth]
  {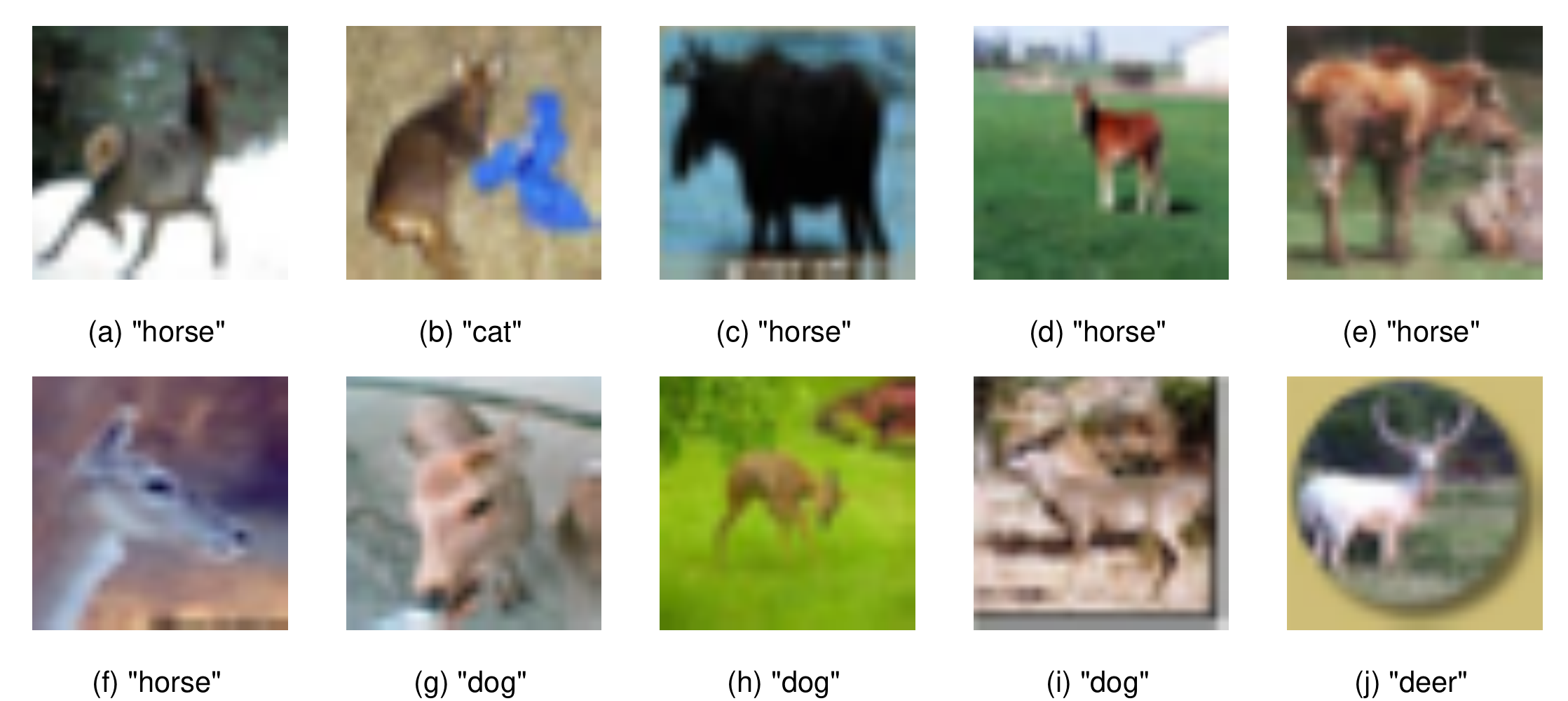}
\caption{Class map of the deer class,
   with the corresponding images.}
\label{fig:cifar_classmap_deer}
\end{figure}
\clearpage

\begin{figure}[!ht]
\vspace{2.3cm}
\centering
\includegraphics[width = 0.6\textwidth]
  {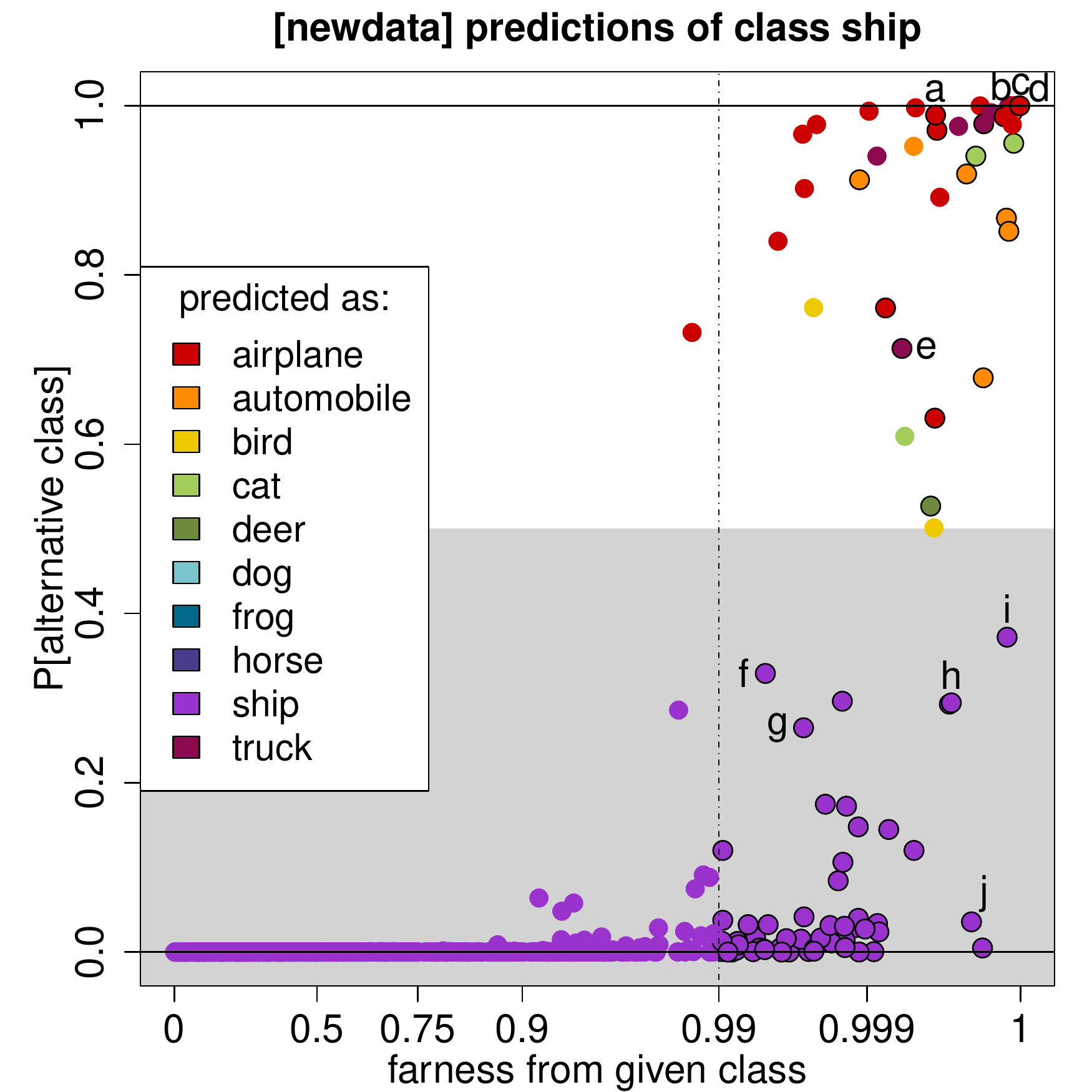}\\
\vspace{0.3cm}
\includegraphics[width = 0.8\textwidth]
  {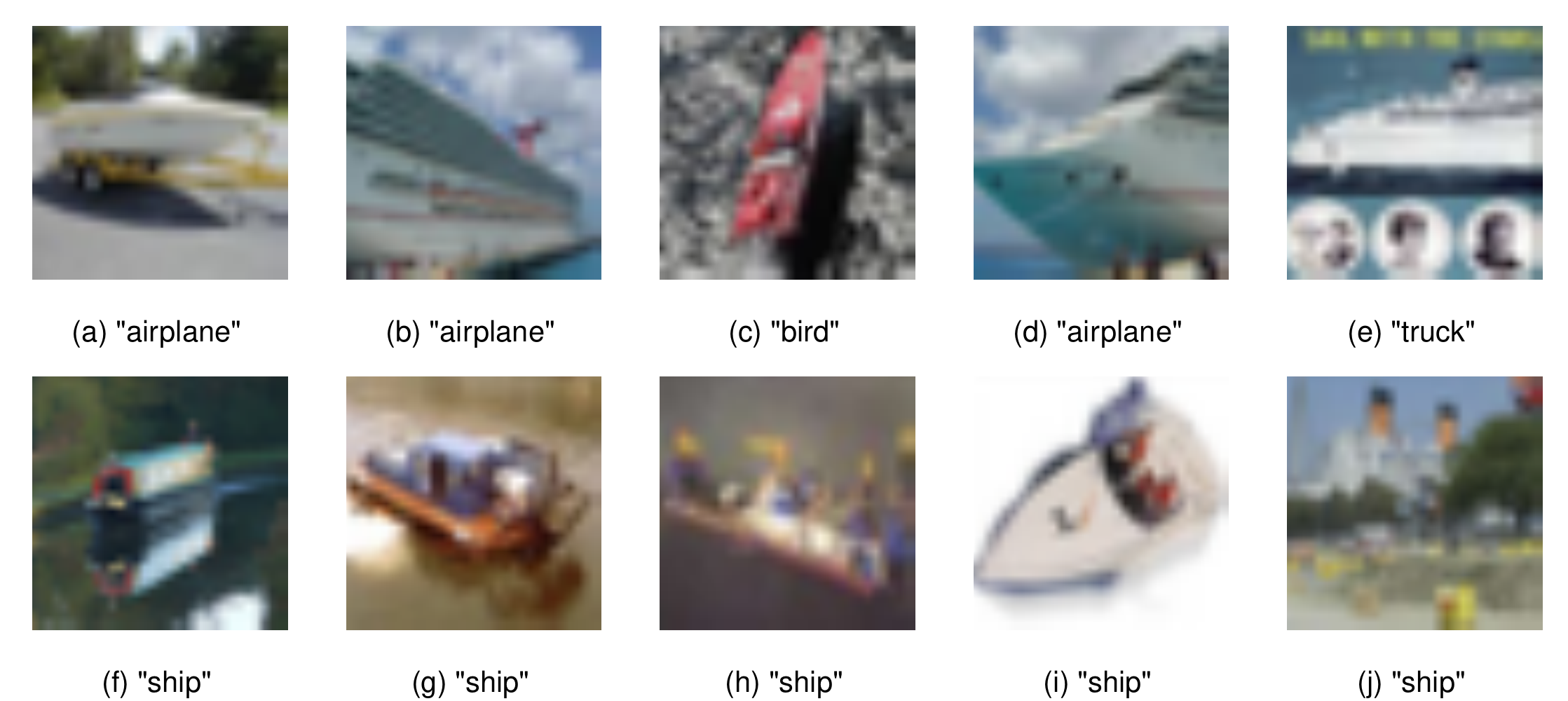}
\caption{Class map of the ship class,
   with the corresponding images.}
\label{fig:cifar_classmap_ship}
\end{figure}
\clearpage

\clearpage
\subsection{Computing farness for tree-based
            classifiers}
\label{suppmat:daisyfarness}

In subsections~\ref{sec:rpart} and \ref{sec:RF} 
we described how to compute interpoint
dissimilarities $d(i,j)$ between any two cases
$i$ and $j$ in the training data, by applying
\texttt{daisy} with variable weights equal to 
each variable's importance as obtained from
the classifier.

The task at hand is to derive a dissimilarity
measure $D(i,g)$ of each training case $i$ 
to every class $g$.
Given that the classes may form disconnected
regions in feature space, the construction
needs to be local rather than global.
For each object $i$ and class $g$ we compute 
$D(i,g)$ as the median of the $k$ smallest 
dissimilarities $d(i,j)$ to all objects $j$ of 
class $g$.
The number $k$ can be chosen by the user.
The default is $k=5$, which worked well in 
a wide range of applications.
For each class $g$ we then divide $D(i,g)$ by
$\median\{D(j,g)\,;\,j\mbox{ belongs to class }g\}$.
This makes the $D(\cdot, g)$ values from all 
classes more comparable to each other.
Finally, we estimate the distribution of the $D(i,g)$
as in the previous section~\ref{suppmat:transfo},
yielding $\farness(i,g)$.

The above formulas can also be used for new
data, such as a test set.
We then start by computing all dissimilarities
$d(i, h)$ where case $i$ belongs to the new 
dataset and $h$ is any case in the training data.
This computation uses the same variable weights
and other parameters as in the training data.
We then compute $D(i, g)$ as the median of the
$k$ smallest dissimilarities $d(i,h)$ to all 
objects $h$ of class $g$ in the training data.
Here $k$ is the same as in the training data.
We then divide $D(i,g)$ by the same denominator
$\median\{D(j,g)\,;\,j\mbox{ belongs to class }g\}$
that was already computed on the training data.
In order to turn the $D(i,g)$ into 
$\farness(i,g)$ we apply the transformation
fitted to the training data in
section~\ref{suppmat:transfo}, that is, we
standardize the $d_i$ with the median and mad
from the training data, then apply the 
Yeo-Johnson transform~\eqref{eq:YJ} with the
same $\lambda$, and then standardize the result
with the same constants as in the training data. 

All of this ensures that the farness of a new 
case in the test set only depends on the training 
data and the new case, and not on other cases
in the test set. In principle, the new dataset
could even consist of a single case.

\clearpage
\subsection{The Titanic test data}
\label{suppmat:titanic}

\begin{figure}[!ht]
\centering
\includegraphics[width=0.7\columnwidth]
   {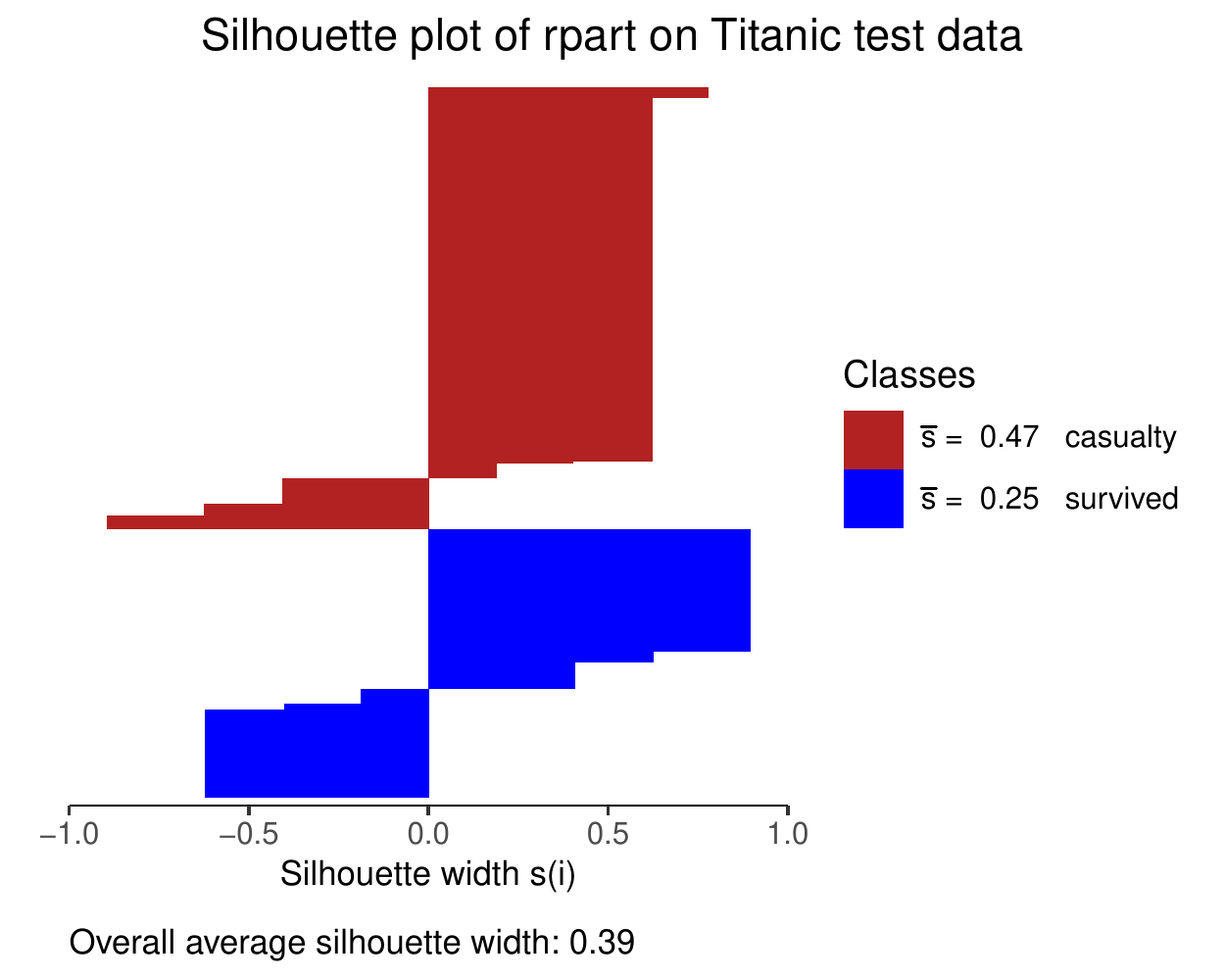}
\caption{Silhouette plot of the classification 
         of the Titanic test data.}
\label{fig:titanic_test_silhouettes}
\end{figure}

We now analyze the Titanic test data. 
The classification tree obtained on the training
data and shown in Figure \ref{fig:titanic_tree}
has an accuracy of about 78\% on the test data,
which is not much lower than the 82\% on the 
training data. 
Figure~\ref{fig:titanic_test_silhouettes} shows
the silhouette plot on the test data. Its 
overall average silhouette width is slightly 
lower than on the training data, so the 
classification is less precise. 
On the other hand, the shape of the 
silhouette plot looks like that of the training 
data, so the classifier behaves in a similar 
fashion here. The class of survivors again 
proved harder to predict than the class of
casualties.

\begin{figure}[!ht]
\includegraphics[width=1.0\columnwidth]
  {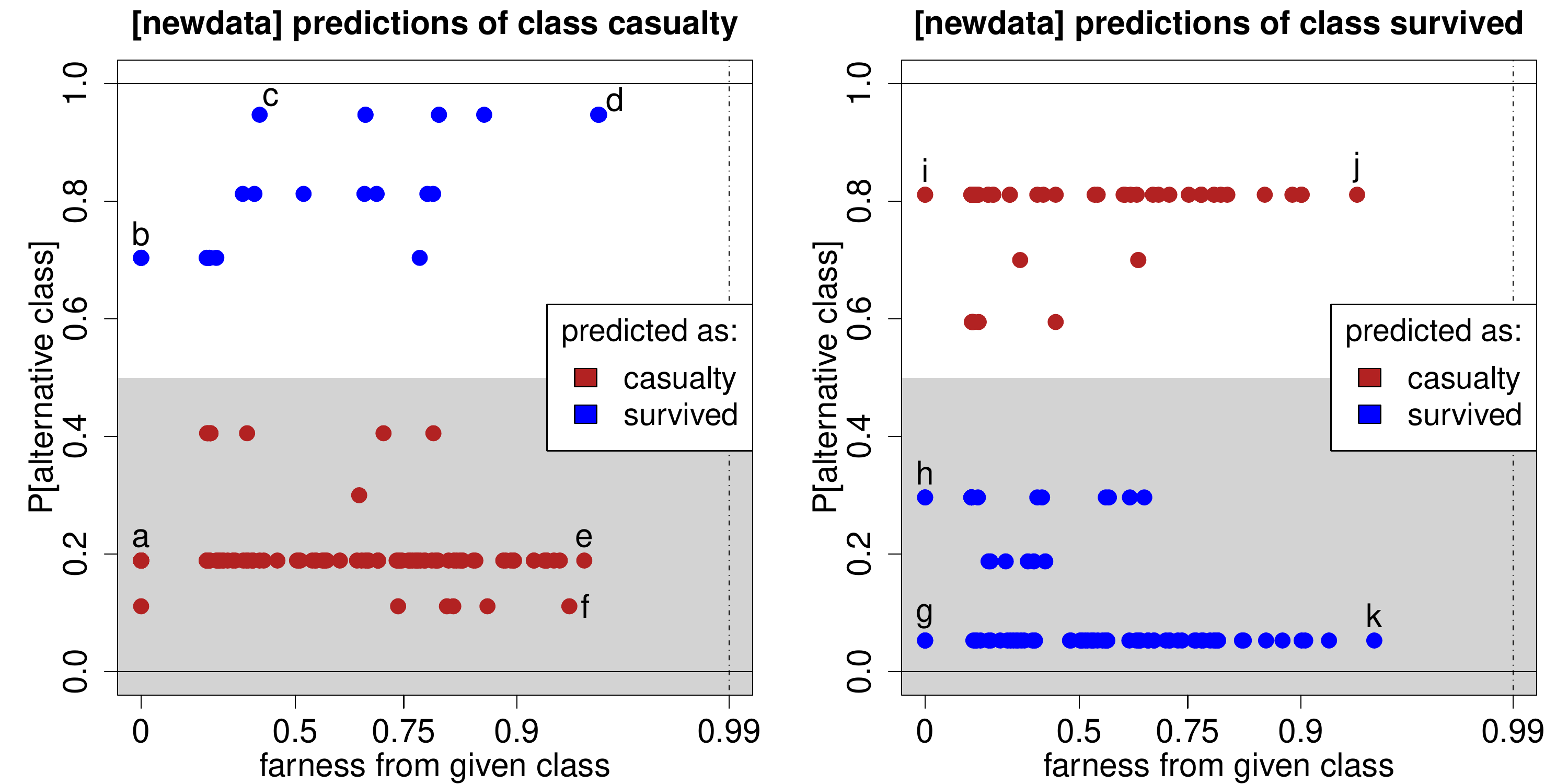}	
\caption{Titanic test data: class maps of 
  classes of casualties (left) and 
	survivors (right).}
\label{fig:titanic_test_classmaps}
\end{figure}

The class maps of both classes are shown in 
Figure~\ref{fig:titanic_test_classmaps}. 
The left panel is from the casualty class.
Passenger \texttt{a} sits well within
the class of casualties and is predicted
as casualty with low PAC, i.e. fairly high 
conviction. It is a male passenger traveling 
in third class without unusual variables.
Case \texttt{b} is a female traveling in 
third class, who paid a low fare and embarked
in Queenstown. She is misclassified
as survivor with mediocre conviction.
Her farness is low since her variables have
typical values.
Case \texttt{c} is also misclassified, but 
with higher conviction than \texttt{b}. 
This is also a female passenger, but
traveling in second class which made her 
survival more likely.
Point \texttt{d} corresponds to a woman 
traveling in first class. This makes her very 
likely to survive, hence her high PAC value.
Within the casualty class, female first 
class travelers were rare.
Finally, passengers \texttt{e} and \texttt{f}
are a husband and wife traveling third class
who paid a low fare, hence they are predicted
as casualties. Their high farness is due
to the fact that they traveled with 9 
parents+children, which is the highest 
number in the test data.\\

The class map of the survivors is shown
in the right hand panel of 
Figure~\ref{fig:titanic_test_classmaps}.
Point \texttt{g} is a female traveler in 
second class without any unusual features,
so the point has a low PAC and farness.
Case \texttt{h} is also a female passenger,
but traveling in third class. This
causes her to be predicted as survivor
with less conviction than \texttt{g}.
Passenger \texttt{i} is a male without
special characteristics, and therefore
predicted as casualty with low farness. 
Point \texttt{j} is also a male passenger,
but he paid a very high fare.
This makes him stand out from the majority
of passengers in the survived class,
explaining his high farness.
Finally, passenger \texttt{k} is a female
traveling in first class. This causes her 
low PAC, that is, she was assigned to the
survivor class with high conviction. 
Her farness is due to paying the highest
fare in the test data.

\clearpage
\subsection{More on the emotion data}
\label{suppmat:emotion}

In subsection \ref{sec:RF} we discussed the 
classes anger and joy. Here we will address the 
two remaining classes.

\begin{figure}
\centering
\includegraphics[width=0.98\columnwidth]
  {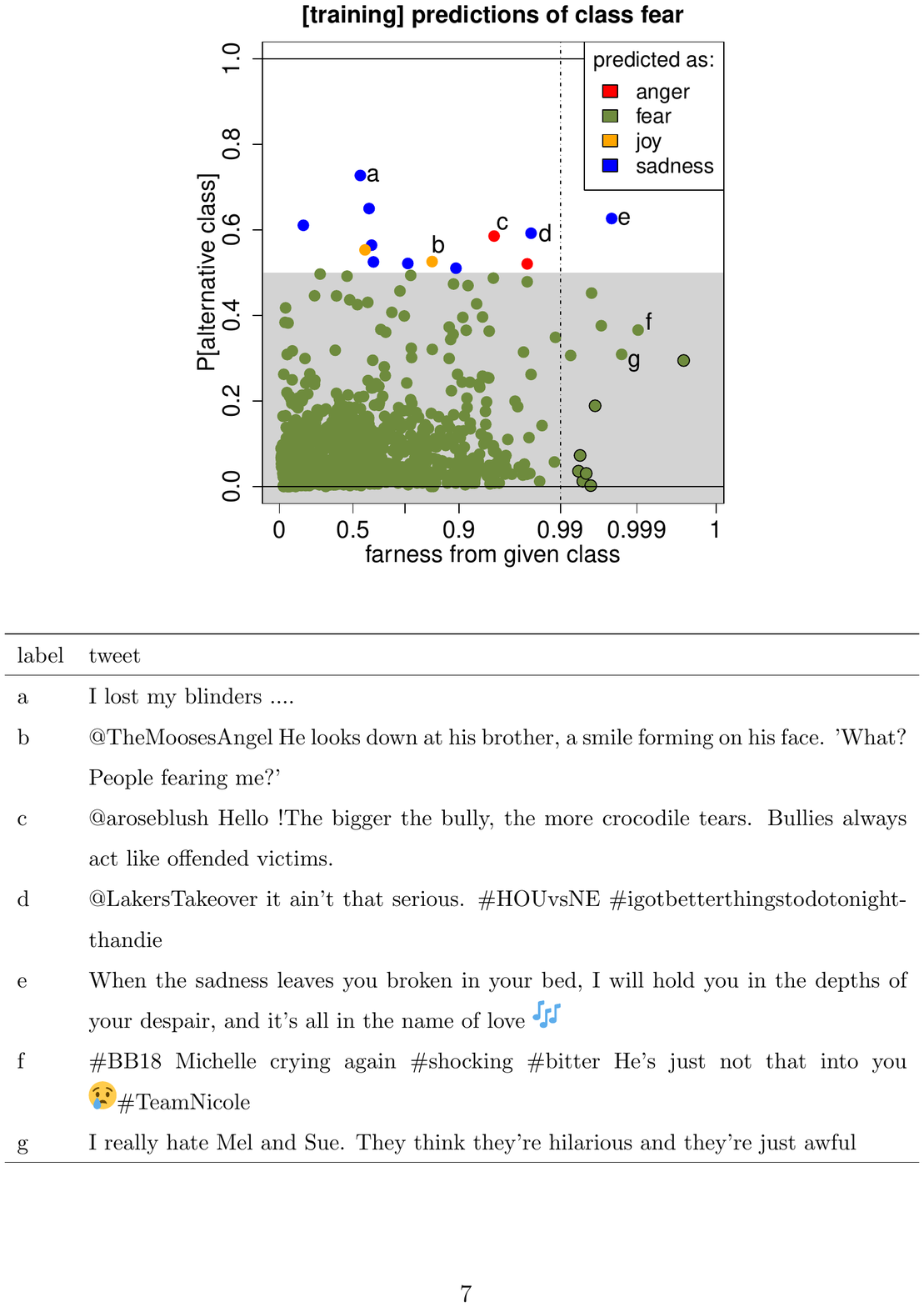}
\caption{Class map of the fear class, with
  the corresponding tweets.}
\label{fig:emotion_classmap_fear}
\end{figure}

Figure~\ref{fig:emotion_classmap_fear} contains
the class map of the fear class.
Most points are classified correctly as fear,
and have unexceptional farness meaning that they
sit well within the class. 
Many of the misclassified points are blue,
indicating some confusion with the sadness class. 
Point \texttt{a} is assigned to sadness due to 
the word `lost'.
Tweet \texttt{b} is predicted as joy due to 
`smile', but with $\PAC(i)$ only slightly above 
0.5 (i.e. not with great conviction) due to 
the word `fearing'.
Tweet \texttt{c} also has a borderline PAC. 
It is predicted as anger, whereas `bully' is 
associated with the fear class.
Tweets \texttt{d} and \texttt{e} are predicted as
sadness due to the words `serious', `sadness' 
and `despair', and it is not clear why they
were labeled as fear in the first place.
The remaining marked points are assigned to 
fear, their given class. 
Tweets \texttt{f} and \texttt{g}
contain the words `shocking' and `awful' 
which are associated with fear.
However, they still have an elevated PAC because
of the words `bitter' and `hilarious'
which are atypical for the fear class. 
They contain several rare n-grams in the 
vocabulary such as `think they' or `do what',
which increased their farness.

\newpage
\clearpage

\begin{figure}
\centering
\includegraphics[width=0.98\columnwidth]
  {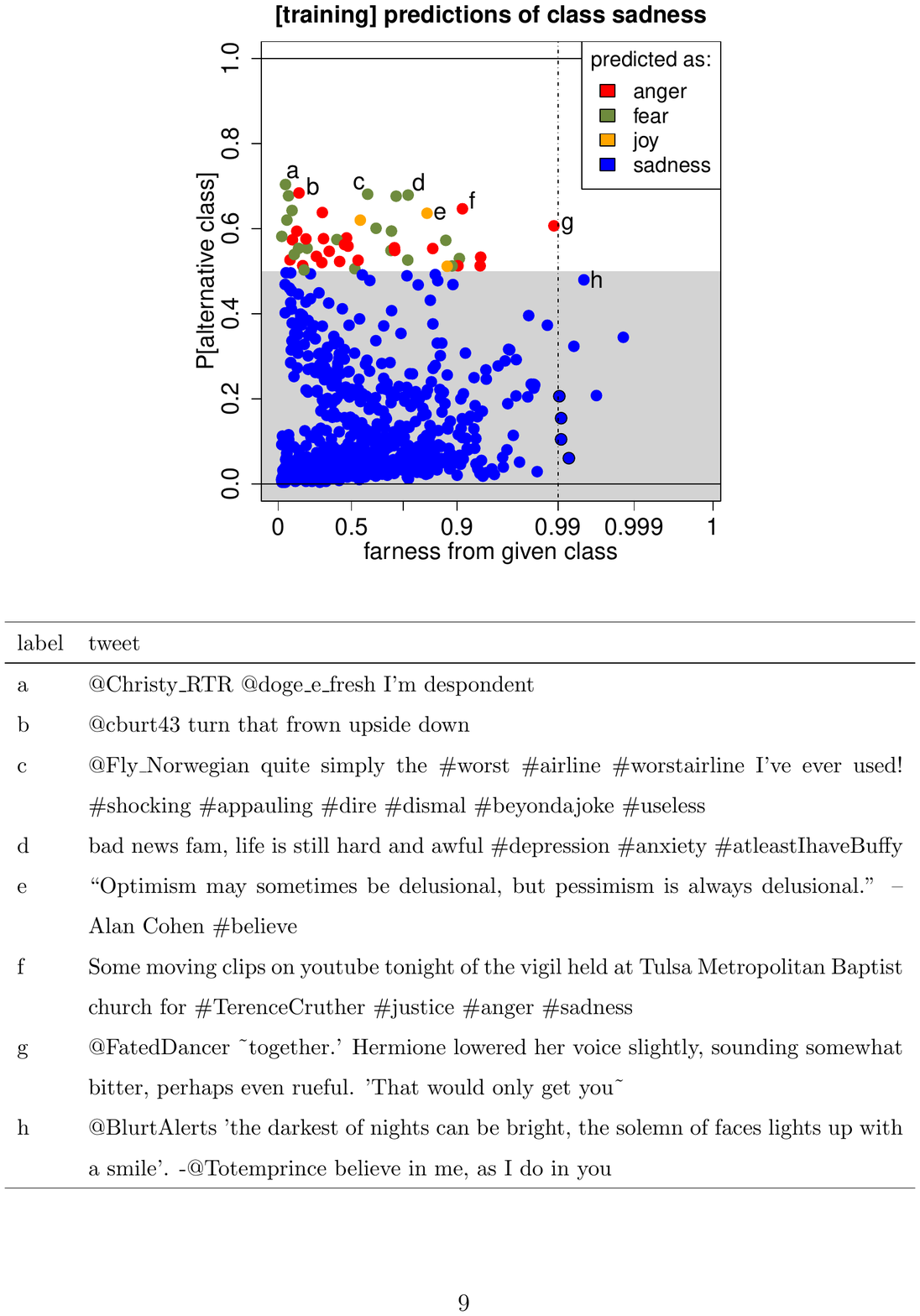}
\caption{Class map of the sadness class, with
  the corresponding tweets.}
\label{fig:emotion_classmap_sadness}
\end{figure}

\newpage
Finally, we discuss the sadness class map presented
in Figure~\ref{fig:emotion_classmap_sadness}.
The majority of the points are blue, so they
were predicted correctly. However, there are
quite a few borderline cases with a PAC value 
somewhat above 0.5\,. 
Most of these are predicted as fear or anger,
emotions that in some sense lie closer to 
sadness than joy does. 
Tweet \texttt{a} is short and does not contain
enough relevant information. 
The word `despondent' was too rare to make the 
vocabulary, so the tweet is predicted in the 
largest class (fear).
Tweet \texttt{b} is a borderline case, as the 
words `frown' and `down' are associated with 
both anger and sadness. 
Tweet \texttt{c} is predicted as fear, but its 
label should probably be anger, rather than 
sadness or fear.
It is predicted as fear due to the word `shocking'.
The words `dismal', `useless', and `worst' point
to anger, but they are quite rare in the data 
and also appear in the fear class.
Tweet \texttt{d} contains `awful' and `anxiety',
causing it to be classified as fear. The 
classification is not with very high conviction 
though, due to the word `depression' pointing
to sadness.
Tweet \texttt{e} is a quote and doesn't have a 
clear emotion connected to it. The classifier 
picks up on the word `optimism' which is 
strongly associated with joy.
Tweets \texttt{f} and \texttt{g} are predicted 
as anger due to the words `anger' and `bitter'. 
Tweet \texttt{h} is a boundary case, containing
words pointing to sadness and others to joy.

\newpage

\end{document}